\documentclass[runningheads]{llncs}

\usepackage{eccv}

\usepackage{eccvabbrv}

\usepackage{graphicx}
\usepackage{booktabs}

\usepackage[accsupp]{axessibility}  

\usepackage{hyperref}

\usepackage{orcidlink}

\begin{document}

\title{RSL-BA: Rolling Shutter Line \\ Bundle Adjustment}

\titlerunning{RSL-BA}

\author{Yongcong Zhang\inst{1,3,*}\orcidlink{0009-0009-1254-6549} \and 
Bangyan Liao\inst{2,*}\orcidlink{0009-0007-7739-4879} \and
Yifei Xue\inst{1,3}\orcidlink{0000-0002-4443-4367} \and
Chen Lu\inst{4}\orcidlink{0009-0003-5779-4673} \and \\
Peidong Liu\inst{2}\orcidlink{0000-0002-9767-6220} \and
Yizhen Lao\inst{1,\dagger}\orcidlink{0000-0002-6284-1724}
}   

\authorrunning{Y. Zhang, B. Liao et al.}


\institute{\textsuperscript{1}Hunan University \quad \textsuperscript{2}Westlake University \quad \textsuperscript{3}DaHe.AI \\ \textsuperscript{4}Dreame Technology Company}
\maketitle
\renewcommand{\thefootnote}{\relax}\footnotetext{\textsuperscript{$\star$} Equal contribution\\  
  \inst{\dagger} Corresponding author: (yizhenlao@hnu.edu.cn) \\
  Project page: \href{https://github.com/zhangtaxue/RSL-BA}{https://github.com/zhangtaxue/RSL-BA}}
\renewcommand{\thefootnote}{\arabic{footnote}}
\setcounter{footnote}{0}

\begin{abstract}
  The line is a prevalent element in man-made environments, inherently encoding spatial structural information, thus making it a more robust choice for feature representation in practical applications. Despite its apparent advantages, previous rolling shutter bundle adjustment (RSBA) methods have only supported sparse feature points, which lack robustness, particularly in degenerate environments. In this paper, we introduce the first rolling shutter line-based bundle adjustment solution, \textit{RSL-BA}. Specifically, we initially establish the rolling shutter camera line projection theory utilizing Plücker line parameterization. Subsequently, we derive a series of reprojection error formulations which are stable and efficient. Finally, we theoretically and experimentally demonstrate that our method can prevent three common degeneracies, one of which is first discovered in this paper. Extensive synthetic and real data experiments demonstrate that our method achieves efficiency and accuracy comparable to existing point-based rolling shutter bundle adjustment solutions.

  \keywords{Rolling Shutter \and Bundle Adjustment}
\end{abstract}

\section{Introduction}
\label{sec:Introduction}

Bundle Adjustment (BA) is a crucial step in multi-view 3D reconstruction, as it jointly optimizes camera poses and scene structure. Although feature point-based BA~\cite{fisher2021colmap, schonberger2016structure, triggs2000bundle} currently dominates the academic landscape, line-based BA has been gaining increasing attention from researchers~\cite{liu20233d, wei2022elsr}. Lines, being one of the most common features in man-made environments, effectively describe the structural information of 3D scenes, in contrast to points which rely solely on local image patches. This structural information provides robust constraints for visual reconstruction~\cite{von2008lsd}, making line a more suitable choice in challenging scenarios.

In parallel with classical bundle adjustment~\cite{triggs2000bundle}, several methods~\cite{lao2021solving,Albl2016} have incorporated the rolling shutter camera model to simultaneously estimate camera poses, structure, and instantaneous motion parameters. However, all existing rolling shutter bundle adjustment (RSBA) methods rely solely on point features, while \textbf{line features have never been addressed in RSBA}. We identify two challenges that hinder the direct combination of line-based BA and the rolling shutter camera model.

\begin{figure}[t]
\centering
\includegraphics[width=1\columnwidth]{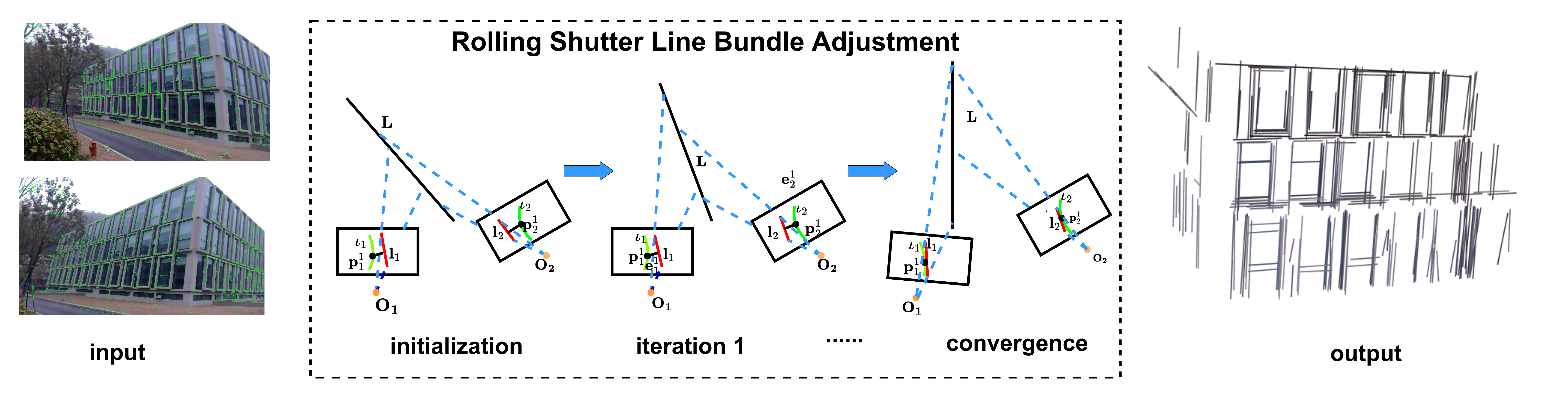}
\caption{The pipeline of proposed \textit{RSL-BA} with some example results. Starting with the input of some RS images (left), the optimization of camera poses and 3D line coordinates is performed using proposed \textit{RSL-BA} (middle), yielding well-reconstructed 3D line segments and accurately estimated camera poses (right).}. 
\label{fig: pipline}
\end{figure}

The first challenge arises from the time-dependent exposure of RS cameras. Unlike CCD cameras and their global shutter (GS) counterparts, RS cameras capture images in a scanline-by-scanline manner. Consequently, as illustrated in Fig.~\ref{fig: pipline}, images taken by moving RS cameras exhibit distortions known as the RS effect. This effect curves the projection of each 3D straight line, making it non-trivial to directly transfer the line-based BA to the rolling shutter camera setting.

The second challenge is associated with degeneracy. As shown in~\cite{Albl2016,zhuang2019learning}, degeneracy is one of the main obstacles in RSBA since iterative optimization can easily be collapsed in some degenerate solutions which are far from the ground truth solutions. Although several strategies have been proposed to mitigate degeneracy in RSBA~\cite{Albl2016,NWMRSBA,lao2018robustified}, a specific strategy to address degeneracy in line-based RSBA is still lacking.

In conclusion, an \textbf{accurate} and \textbf{robust} solution to line-based RSBA is still missing. Such a method would be vital in the potential widespread deployment of 3D vision with RS imaging systems.

\subsection{Related Works}
\label{sec:Related Works}

\textbf{Video-based RSBA.}The assumption of smooth continuous trajectories is widely employed for RS video inputs to reduce the optimization parameter space and enhance algorithm robustness. In~\cite{Hedborg2012}, Hedborg \textit{et al.} present an RSBA algorithm that interpolates the motion between consecutive frames. Zhuang \textit{et al.}~\cite{zhuang2017rolling} further propose an optical flow-based RSBA, which are developed to recover the relative pose of an RS camera that undergoes constant velocity and acceleration motion, respectively. Following this assumption, a spline-based camera trajectory motion model is proposed by~\cite{patron2015spline}. 

\noindent \textbf{Unordered RSBA.} As the unordered image set is the standard input for SfM. Albl \textit{et al.} first addresses this setting by explicitly velocity modeling and optimization. Besides, a planar degeneracy configuration has also been disclosed in ~\cite{Albl2016}. While in~\cite{lao2018robustified}, Lao~\textit{et al.} propose a camera-based RSBA to simulate the actual camera projection, exhibiting the degeneracy resilience ability. In \cite{ito2016self}, 
Ito \textit{et al.} establish the equivalence between self-calibrated SfM and RSSFM based on the pure rotation instantaneous motion model and affine camera assumption, while the work of~\cite{lao2021solving} draws the equivalence between RSSfM and non-rigid SfM. Recently, Liao \textit{et al.} proposed two techniques to boost the accuracy and robustness of RSBA in \cite{NWMRSBA}.

\noindent \textbf{Line-based GSBA.} In~\cite{taylor1995structure, bartoli2005structure}, the authors propose a comprehensive line-based SFM pipeline, while~\cite{holynski2020reducing, micusik2017structure} introduce line-based incremental SFM frameworks. Wei \textit{et al.} in~\cite{wei2022elsr} employs planes and points to guide matching, achieving efficient line reconstruction. Recently, Lui \textit{et al.} introduced LIMAP in~\cite{liu20233d}, where the authors devised various strategies and algorithms to achieve efficient and precise line reconstruction. In~\cite{zuo2017robust, pumarola2017pl, lim2022uv}, hybrid strategies combining lines and feature points are utilized to enhance the accuracy and robustness of SLAM.

To this end, we propose a novel \textit{RSL-BA} algorithm to solve the line-based RSBA problem. Specifically, we first establish the rolling shutter line projection theory using the Plücker line parameterization and subsequently derive a series of reprojection error formulations. We claim that the proposed reprojection error formulations are efficient and exhibit the degeneration-resistant ability. We also provide complete degeneration-resistant proof of our proposed error formulation. Our contributions are summarized as follows:
\begin{itemize}
    \item To the best of our knowledge, this is the first line-based RSBA solution, which serves as the foundation for line-based or point-line-hybrid RS-SfM and RS-SLAM.
    \item we theoretically and experimentally demonstrate that the proposed \textit{RSL-BA} can prevent three common degeneracies in RSBA, one of the degenerate cases is first discovered in this paper.
    \item The extensive evaluations in both synthetic and real datasets exhibit the comparable efficiency and accuracy of the proposed method over previous works. 
\end{itemize}

\section{Background}
\label{sec:Background}
In this section, we review the rolling shutter projection model in Sec.~\ref{subsec:rs_model} and the parameterization of 3D lines in Sec.~\ref{subsec:3d_line}. 
\subsection{Rolling Shutter Camera Projection Mode}
\label{subsec:rs_model}
Let $\mathbf{R}(v) \in \mathbf{SO}({3})$ and $\mathbf{t}(v) \in \mathbb{R}^{3}$ represent the camera rotation and translation, respectively, when the row index $v$ of measurement is acquired. Denote the intrinsic camera parameters as $\mathbf{K}$, and the initial rotation and translation of the camera at $v=0$ as $\mathbf{R}_{0}$ and $\mathbf{t}_{0}$, respectively. The projection matrix of the global shutter camera is defined as $ \mathbf{P}_0 = \mathbf{K}[\mathbf{R}_0, \mathbf{t}_0]$. Given the assumption of constant camera motion during frame capture, which is commonly made in RS 3D vision tasks~\cite{Dai,Albl2016,lao2021solving}, we can model the instantaneous motion as:
\begin{equation}\label{equation:motion_linear}
    \begin{aligned}
        \mathbf{R}(v) = (\mathbf{I} + [\boldsymbol{\omega}]_{\times} v)\mathbf{R}_{0},\qquad\mathbf{t}(v) = \mathbf{t}_{0} + \mathbf{d} v,     
    \end{aligned}
\end{equation}
 where $\mathbf{d} = [d_{x}, d_{y}, d_{z}]^{\top}$ is the translational velocity vector and $\boldsymbol{\omega} = [\omega_{x},\omega_{y},\omega_{z}]^{\top}$ is the rotational velocity vector , and $[\boldsymbol{\omega}]_{\times}$ represents the skew-symmetric matrix of vector $\boldsymbol{\omega}$. Then, the projection matrix of the camera under the RS model can be defined as:
\begin{equation}\label{equ:rs_projection_matrix}
    \begin{aligned} 
        \mathbf{P}_{v} = \mathbf{K}[\mathbf{R}_v, \mathbf{t}_v]  &= \mathbf{P}_{0}+v\mathbf{Q}\\
        \mathbf{P}_{0} = \mathbf{K}[\mathbf{R}_0, \mathbf{t}_0] \quad &\mathbf{Q} = \mathbf{K}[[\boldsymbol{\omega}]_{\times}\mathbf{R}_{0}, \mathbf{d}].
    \end{aligned}
\end{equation}
\subsection{3D Line Segments Parameterization}
\label{subsec:3d_line}
We employ the Plücker matrix to parameterize the 3D line~\cite{Hartley2003}. Specifically, given two homogeneous 3D points ${P_{a}}=[a_1,a_2,a_3,1]^{\top}$ and ${P_b}=[b_1,b_2,b_3,1]^{\top}$. The Plücker matrix $\mathbf{L}$ can be defined as:
\begin{equation}
\begin{split}
\mathbf{L} = {P_{a}P_{b}^{\top}-P_{b}P_{a}^{\top}} \\ 
= \begin{bmatrix}
0 & a_1b_2-a_2b_1 & a_1b_3-a_3b_1 & a_1-b_1 \\
a_2b_1-a_1b_2 & 0 & a_2b_3-a_3b_2 & a_2-b_2 \\
a_3b_1-a_1b_3 & a_3b_2-a_2b_3 & 0 & a_3-b_3 \\
b_1-a_1 & b_2-a_2 & b_3-a_3& 0
\end{bmatrix} 
= \begin{bmatrix}
0 & l_{12} & l_{13} & l_{14} \\
-l_{12} & 0 & l_{23} & l_{24} \\
-l_{13} & -l_{23} & 0 & l_{34} \\
-l_{14} & -l_{24} & -l_{34}& 0
\end{bmatrix},
\end{split}
\end{equation}
where the $[l_{14}, l_{24}, l_{34}]^{\top}$ represents the direction of this straight line while the $[-l_{23}, l_{13}, -l_{12}]^{\top}$ represents the normal direction of this line. The projection of 3D  Plücker matrix is the skew-symmetric matrix of a 2D normalized line. It can be related to each other as:
\begin{equation}\label{equ:gs_line_projection}
    \begin{aligned} 
        [\mathbf{l}_{gs}]_{\times} = [\mathbf{R}, \mathbf{t}] \mathbf{L} [\mathbf{R}, \mathbf{t}]^{\top}
    \end{aligned}
\end{equation}

\section{Methodology}
\label{sec:Methodology}
In Sec.~\ref{subsec:Rolling Shutter Line Projection Formulation}, we initially establish the rolling shutter projection equations using the Plücker line parameterization. These equations establish a connection between the 3D lines and the curve parameters on the image plane. Then in Sec.~\ref{subsec:Error_Definition}, we present a series of line-based reprojection error equations. On top of the error definition, we can formulate the entire rolling shutter line bundle adjustment problem in Sec.~\ref{subsec:Rolling Shutter Line Bundle Adjustment}. Finally, in Sec.~\ref{subsec:Degeneracies Analysis}, we demonstrate the degeneracy resilience capability of the proposed error equations.

\subsection{Rolling Shutter Line Projection Formulation}
\label{subsec:Rolling Shutter Line Projection Formulation}
As previously explained, the curved rolling shutter projection of 3D lines occurs because of the camera's nonlinear motion. Using the Plücker parameterization method, we can derive explicit expressions for the curve parameters.

Firstly, we construct the skew-symmetric matrix $[\mathbf{l}_{rs}]_{\times}$ of the instantaneous projection line by combining Eq.~(\ref{equ:rs_projection_matrix}) and Eq.~(\ref{equ:gs_line_projection}):
\begin{equation}
\begin{aligned} 
{[\mathbf{l}_{rs}]_{\times}} &= \mathbf{P}_{v}\mathbf{L}\mathbf{P}_{v}^{\top} \\
&= \mathbf{P}_{0}\mathbf{L}\mathbf{P}_{0}^{\top} + v (\mathbf{P}_{0}\mathbf{L}\mathbf{Q}^{\top}+\mathbf{Q}\mathbf{L}\mathbf{P}_{0}^{\top}) + v^{2}\mathbf{Q}\mathbf{L}\mathbf{Q}^{\top}\\
&= \mathbf{A}_1 + v \mathbf{A}_2 + v^{2} \mathbf{A}_3 
\end{aligned}
\end{equation}

Secondly, as $\mathbf{l}_{rs} = [l_1,l_2,l_3]^{\top}$ and $[\mathbf{l}_{rs}]_{\times} =  \mathbf{A}_1 + v \mathbf{A}_2 + v^{2} \mathbf{A}_3 $, we can determine the instantaneous projected line parameter as follows:
\begin{equation}
\left\{
\begin{array}{l}
l_1 = \mathbf{A}_{1}^{32}+v\mathbf{A}_{2}^{32}+v^2\mathbf{A}_{3}^{32}\\
l_2 = \mathbf{A}_{1}^{13}+v\mathbf{A}_{2}^{13}+v^2\mathbf{A}_{3}^{13}\\
l_3 = \mathbf{A}_{1}^{21}+v\mathbf{A}_{2}^{21}+v^2\mathbf{A}_{3}^{21}
\end{array} \right. 
\label{eq:projected staight line definition}
\end{equation}

Ultimately, we note that the instantaneous projected line parameter fulfills this $l_{1}u+l_{2}v+l_{3}=0$ equation, which we can expand to obtain the final curve equation, denoted as $\boldsymbol{\iota}$:
\begin{equation}
\begin{split}
\boldsymbol{\iota} : \mathbf{A}_{3}^{13}v^{3}+\mathbf{A}_{3}^{32}uv^{2}+(\mathbf{A}_{2}^{13}+\mathbf{A}_{3}^{21})v^{2}+\mathbf{A}_{2}^{32}uv
\\+(\mathbf{A}_{1}^{13}+\mathbf{A}_{2}^{21})v+\mathbf{A}_{1}^{32}u+\mathbf{A}_{1}^{21}=0
\end{split}
\label{eq:projected curve}
\end{equation}

Our conclusion aligns with that of Lao \textit{et al.}~\cite{lao2018robust}, yielding an identical polynomial curve degree. Nevertheless, our utilization of a 4-DoF orthogonal representation representation for 3D lines has distinct benefits when compared to Lao's 5-DoF over-parameterized representation. These advantages include the capability to conduct unconstrained optimization on spatial lines directly.

\textit{Remark.} The geometric interpretation underlying this derivation is quite straightforward. Given the parameters of a spatial line and the current camera pose, if point (u, v) lies on the curve obtained by projecting this spatial line, then when the RS camera scans to the v-th row, the projection of the spatial line onto the image plane follows Eq.~(\ref{eq:projected staight line definition}). Therefore, it is evident that point (u, v) must be located on this line. Based on this reasoning, we can deduce Eq.~(\ref{eq:projected curve}).

\subsection{Line-based Reprojection Errors}
\label{subsec:Error_Definition}
\begin{figure}
\centering
\includegraphics[width=1\columnwidth]{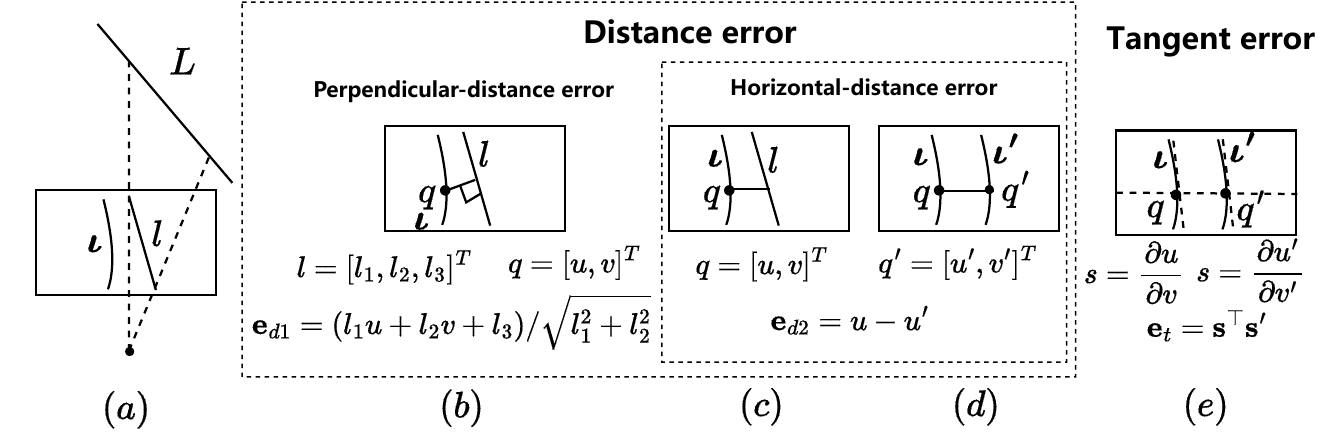}
\caption{\textbf{The illustration of proposed line-based reprojection errors given the spatial line ${L}$ and observed curve $\boldsymbol{\iota}$.}
Instead of directly measuring the distance between the observed curve and the projected curve, we split each curve into multiple points $q$, and aggregate the distance errors. Specifically, we first utilize the virtual projected line $l$ when observing it, as shown in (a). To measure the distance between each point $q$ and the corresponding virtual lines, we can use the perpendicular distance (b), the horizontal distance (c,d) or the tangent distance (e) as our basic metrics.
}
\label{fig: line_projection}
\end{figure}
As claimed in Sec.~\ref{subsec:Rolling Shutter Line Projection Formulation}, a 3D straight line will become a polynomial curve on the captured image. In this section, we will establish a series of re-projection errors which are stable and robust. We first leverage the similar intuition in Sec.~\ref{subsec:Rolling Shutter Line Projection Formulation} to obtain the distance error representation. As shown in Fig.~\ref{fig: line_projection}(a), Let there be a point $\mathbf{q}=[u, v]^{\top}$ on the observed curve $\boldsymbol{\iota}$, then the pose of the camera when exposing that point is $\mathbf{R}_v$, $\mathbf{t}_v$, then projecting the 3D line $\mathbf{L}$ onto the image plane, this results in the virtual 2D line $\mathbf{l} = [l_1, l_2, l_3]^{\top}$. Under the condition of ground truth, this virtual line must theoretically intersect with that point $\mathbf{q}$. This constraint directly leads us to the two variants of distance error, which we have defined below.

\subsubsection{Perpendicular Distance Error $\mathbf{e}_{d1}$}. The first distance error formulation, named perpendicular distance error, measures the shortest distance from point $\mathbf{q}$ to the line $\mathbf{l}$, as shown in Fig.~\ref{fig: line_projection}(b). The error is defined as:
\begin{equation}
\mathbf{e}_{d1} = \frac{l_{1}u+l_{2}v+l_{3}}{\sqrt{l_{1}^{2}+l_{2}^{2}}},
\end{equation}
where the line parameters $[l_1,l_2,l_3]$ can be calculated through Eq.~(\ref{eq:projected staight line definition}).

\subsubsection{Horizontal Distance Error $\mathbf{e}_{d2}$}. The second distance error formulation, named horizontal distance error, measures the u-axis distance from point $\mathbf{q}$ to the line $\mathbf{l}$, as shown in Fig.~\ref{fig: line_projection}(c). The error can be defined as:
\begin{equation}
\mathbf{e}_{d2} = u-u',
\end{equation}
where the corresponding $u'$ coordinate of virtual line $\mathbf{l}$ at the $v$-th row can be calculated as:
\begin{equation}
u'=-\frac{l_{2}v+l_{3}}{l_{1}}, 
\end{equation}
As shown in Fig.~\ref{fig: line_projection}(d), the $e_{d2}$ distance error is equivalent to the u-axis distance from the point to the projection curve.

\subsubsection{Tangent Error $\mathbf{e}_{t}$}. While distance error provides an elegant constraint, it is insufficient to rely solely on distance error in certain scenarios. To further enhance the robustness, we introduce a second type of tangent error. As shown in Fig.~\ref{fig: line_projection}(e), we constrain the tangent of the projected curve $\boldsymbol{\iota'}$ to match the observed tangent direction. The error can be defined as:
\begin{equation}
\mathbf{e}_{t} = \mathbf{s}^{\top} \mathbf{s'},
\end{equation}
where $\mathbf{s}$ represents the tangent direction observation. The $\mathbf{s'}$ can be calculated analytically following the implicit function theorem as:

\begin{equation}
\mathbf{s'} = [\frac{1}{\sqrt{1+s^2} }, \frac{s}{\sqrt{1+s^2}}], \quad
s=\frac{\partial u}{\partial v} = - \frac{\partial \boldsymbol{\iota} / \partial v}{\partial \boldsymbol{\iota} / \partial u}
\end{equation}

\subsection{Rolling Shutter Line Bundle Adjustment}
\label{subsec:Rolling Shutter Line Bundle Adjustment}
The non-linear least squares solvers are used to find an optimal solution $\boldsymbol{\theta}^{*}$ including camera poses $\mathbf{R}^{*},\mathbf{t}^{*}$, instantaneous motion $\boldsymbol{\omega}^{*},\mathbf{d}^{*}$ and 3D lines $\boldsymbol{\tau}^{*}$ by minimizing the reprojection error $\mathbf{e}_{i}^{j}$ from line $i$ to camera $j$ over all the camera index in set $\mathcal{F}$ and corresponding observed 3D lines index in subset $\mathcal{P}_j$:
\begin{equation}
    \begin{aligned}
        \boldsymbol{\theta}^* =  \left \{ \boldsymbol{\tau}^{*} , \mathbf{R}^{*},\mathbf{t}^{*}, \boldsymbol{\omega}^{*},\mathbf{d}^{*} \right \} = \mathop{\arg\min}_{\boldsymbol{\theta} }  \sum_{j\in \mathcal{F}}\sum_{i\in \mathcal{P}_j }  \left \| \mathbf{e}_{i}^{j} \right \|^{2}_{2}.
    \end{aligned}
    \label{equation:cost_function}
\end{equation}
On top of errors proposed in Sec.~\ref{subsec:Error_Definition} , we provide two variants of reprojection error:
\begin{equation}
\mathbf{e}_1 = \mathbf{e}_{d1} + \lambda_1 \mathbf{e}_{t}, \quad
\mathbf{e}_2 = \mathbf{e}_{d2} + \lambda_2 \mathbf{e}_{t}, 
\end{equation}
where ${\lambda}_1,{\lambda}_2$ are the relative weight. We have experimentally validated the optimality of these two error variants in Sec.~\ref{sec: Synthetic Experiments}. After definition, this non-linear least square problem can be iteratively solved through Levenberg-Marquardt~\cite{more2006levenberg} method.

Besides, the full analytical Jacobian has been derived in the supplementary material.

\subsection{Degeneracy Analysis}
\label{subsec:Degeneracies Analysis}
In this section, we conducted theoretical analysis on the resilience of \textit{RSL-BA} to three types of degeneracy scenarios. These three degeneracy scenarios include plane degeneracy case~\cite{Albl2016},  2-views pure translation case~\cite{zhuang2019learning}, and X-Y pure translation degeneracy case disclosed by us. We assume the positive direction of X Y Z axis is right down forward, respectively.

\subsubsection{\textcircled{1} Resistance to Plane Degeneracy~\cite{Albl2016}.}
\label{sec: plane degeneracy}

We first briefly describe the conditions under which the plane degeneration occurs: when the camera's y-axis is parallel, if a rotational velocity $\boldsymbol{\omega} = [-1, 0,  0]^{\top}$ is given to the camera, then the reconstructed 3D points will be compressed onto a plane perpendicular to the y-axis, and satisfy that the reprojection error is 0.

For line features, we also assume $\mathbf{R}_{0}=\mathbf{I}$, $\mathbf{t}_{0}=\mathbf{0}$, $\boldsymbol{\omega} = [-1, 0, 0]^{\top}$, $\mathbf{d}={0}$, Let there be two points, $\mathbf{P}_a$ and $\mathbf{P}_b$, on the spatial line $\textbf{L}$. $\mathbf{P}_a=[a_1, a_2, a_3, 1]^{\top}$ and $\mathbf{P}_b=[b_1, b_2, b_3, 1]^{\top}$, When the space is compressed into a plane, the new coordinates of $\mathbf{P}_a$ and $\mathbf{P}_b$ are $\mathbf{P}_a'=[a_1, 0, a_3, 1]^{\top}$ and $\mathbf{P}_b'=[b_1, 0, b_3, 1]^{\top}$, respectively. By referring to Sec. \ref{subsec:Rolling Shutter Line Projection Formulation}, we can solve for the projection parameters of the two-dimensional curve at this moment. The parameters of the curve at this time are:
\begin{equation}
\left\{
\begin{array}{l}
\mathbf{A}_{3}^{13} = \mathbf{A}_{3}^{32} = \mathbf{A}_{2}^{13} = \mathbf{A}_{3}^{21} = \mathbf{A}_{2}^{32} = \mathbf{A}_{1}^{32} = \mathbf{A}_{1}^{21} = 0 \\
\mathbf{A}_{1}^{13} = a_{3}b_{1}-a_{1}b_{3} \\
\mathbf{A}_{2}^{21} = a_{1}b_{3}-a_{3}b_{1} 
\end{array} \right. 
\end{equation}

Substituting the parameters into the curve expression Eq. (\ref{eq:projected curve}), we obtain $0 = 0$,  indicating that u and v are unrestricted. Under the influence of $\boldsymbol{\omega} = [-1 \quad 0 \quad 0]^{\top}$,
the camera's imaging rows are always coplanar with the line, making the distance from any point on the image to the line zero and satisfying the point-to-line distance error function (Sec.~\ref{subsec:Error_Definition}). However, the slope constraint cannot be met. Using this projection equation to calculate the tangent direction of any point on the curve results in indeterminate forms $[\frac{0}{0}, \frac{0}{0}]$, which do not match the measured tangent direction. Therefore, using the tangent direction of points on the curve as the error function prevents this degeneration.
As verified in Fig~\ref{fig: proof_degeneracy_experiment}, our proposed \textit{RSL-BA} can suppress plane degeneration, whereas conventional rolling shutter point-based bundle adjustment method NMRSBA~\cite{Albl2016} converges to a degenerate solution.

\subsubsection{\textcircled{2} Resistance to 2-views Pure Translation Degeneracy~\cite{zhuang2019learning}.}
\label{sec: 2-views Pure Translation}

We first briefly describe the conditions under which this degeneration occurs: when the camera moves in a straight line to take two pictures, assigning any translational velocity to the camera for both shots, it is possible to find new 3D points that satisfy the condition that the reprojection error of the points under this velocity is zero.

Let's assume the real pose of the camera before the degeneration is: $\mathbf{R}_{0}=\mathbf{I}$, $\mathbf{t}_{0}=\mathbf{0}$, $\boldsymbol{\omega} = [0, 0, 0]^{\top}$, $\mathbf{d}=[d_1, d_2, d_3]^{\top}$, $\mathbf{L}=(l_{12},l_{13},l_{23},l_{14},l_{24},l_{34})$
In this case, substitute these parameters into Eq.~(\ref{eq:projected curve}) we can get curve expression:
\begin{equation}
\begin{array}{l}
(l_{14}d_{3}-l_{34}d_{1})v^2+(l_{34}d_{2}-l_{24}d_{3})uv
+(l_{24}d_{1}-l_{14}d_{2}+l_{13})v-l_{23}u-l_{12} = 0
\end{array} 
\end{equation}
After the degeneration occurs, $\mathbf{R}_{0}'=\mathbf{I}$, $\mathbf{t}_{0}'=\mathbf{0}$, $\boldsymbol{\omega}' = [0 \quad 0\quad 0]^{\top}$, $\mathbf{d}'=r[d_1, d_2, d_3]$, $\mathbf{L}=(l'_{12},l'_{13},l'_{23},l'_{14},l'_{24},l'_{34})$. In this case, the curve expression is:
\begin{equation}
\begin{array}{l}
r(l'_{14}d_{3}-l'_{34}d_{1})v^2+r(l'_{34}d_{2}-l'_{24}d_{3})uv
+(r(l'_{24}d_{1}-l'_{14}d_{2})+l'_{13})v-l'_{23}u-l'_{12} = 0
\end{array} 
\end{equation}

To make the reprojection error zero, the curve expressions before and after degeneration must be the same, meaning the corresponding coefficients are in proportion.
\begin{equation}
\left\{
\begin{array}{l}
l_{14}d_{3}-l_{34}d_{1} = sr(l'_{14}d_{3}-l'_{34}d_{1}) \\
l_{34}d_{2}-l_{24}d_{3} = sr(l'_{34}d_{2}-l'_{24}d_{3}) \\
l_{24}d_{1}-l_{14}d_{2}+l_{13} = sr(l'_{24}d_{1}-l'_{14}d_{2})+sl'_{13} \\
l_{23} = sl'_{23} \\
l_{12} = sl'_{12} 
\end{array} \right. 
\end{equation}
Substituting the first and second equations into the third equation yields $l_{13}=sl'_{13}$. Combining this with: 
\begin{equation}
\left\{
\begin{array}{l}
l_{12}l_{34}-l_{13}l_{24}+l_{14}l_{23} = 0 \\
l'_{12}l'_{34}-l'_{13}l'_{24}+l'_{14}l'_{23} = 0
\end{array} \right. 
\end{equation}
we obtain:
\begin{equation}
\begin{array}{l}
l'_{12} = \frac{l'_{12}}{s} , \quad l'_{13} = \frac{l_{13}}{s} , \quad l'_{23} = \frac{l_{23}}{s} ,\quad 
l'_{14} = \frac{l_{14}}{sr} , \quad l'_{24} = \frac{l_{24}}{sr} , \quad l'_{34} = \frac{l_{34}}{sr} 
\end{array} 
\end{equation}

\begin{figure}[t]
\centering
\includegraphics[width=0.8\columnwidth]{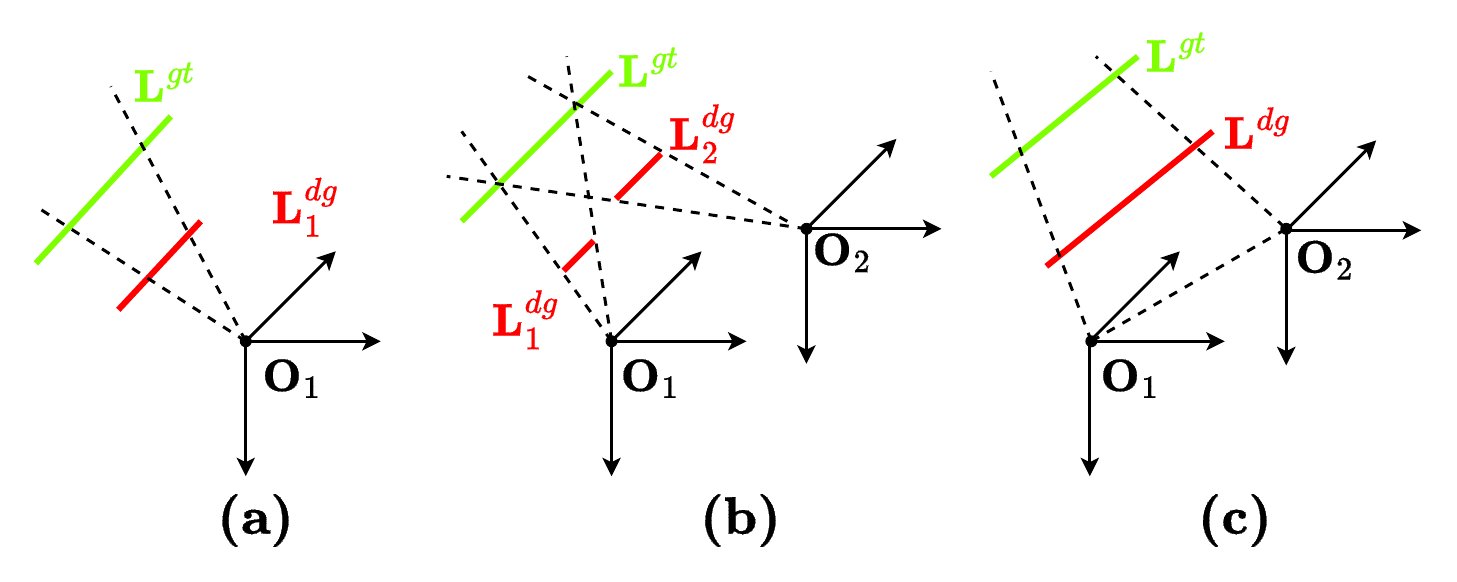}
\caption{
The illustration of degeneracy resistance of our proposed method in 2-views pure translation degeneracy. (a) Degeneration in single-view scenarios. (b) Suppression of degeneration using line features in dual-view scenarios. (c) Inability of line features to suppress degeneration in special dual-view scenarios}
\label{fig: proof_2_views_degeneracy}
\end{figure}

Note that the direction and normal direction of the line remain unchanged, indicating that the degenerated line stays in the original plane and parallel to the original line, as shown in Fig.~\ref{fig: proof_2_views_degeneracy}(1), where $\mathbf{L}^{gt}$ is the original line, and $\mathbf{L}_{1}^{dg}$ is the degenerated line. In the dual-frame scenario (Fig.\ref{fig: proof_2_views_degeneracy}(2)), when the plane $\mathbf{O}_{1}\mathbf{L}^{gt}$ is not coplanar with the plane $\mathbf{O}_{2}\mathbf{L}^{gt}$, their degenerated states $\mathbf{L}_{1}^{dg}$ and $\mathbf{L}_{2}^{gt}$ cannot coincide (except at $\mathbf{L}^{gt}$). In this case, line constraints can prevent degeneration. However, when $\mathbf{L}^{gt}$, $\mathbf{O}_{1}$, and $\mathbf{O}_{2}$ are coplanar (Fig.\ref{fig: proof_2_views_degeneracy}(3)), line constraints cannot suppress degeneration, though this is a very special condition requiring all lines to meet this criterion. This demonstrates that line constraints are more effective than point constraints in preventing degeneration.

\subsubsection{\textcircled{3} Resistance to X-Y Pure Translation Degeneracy (our new founding).}
\label{sec: X-Y Pure Translation Degeneracy}
\begin{figure}
\centering
\includegraphics[width=0.8\columnwidth]{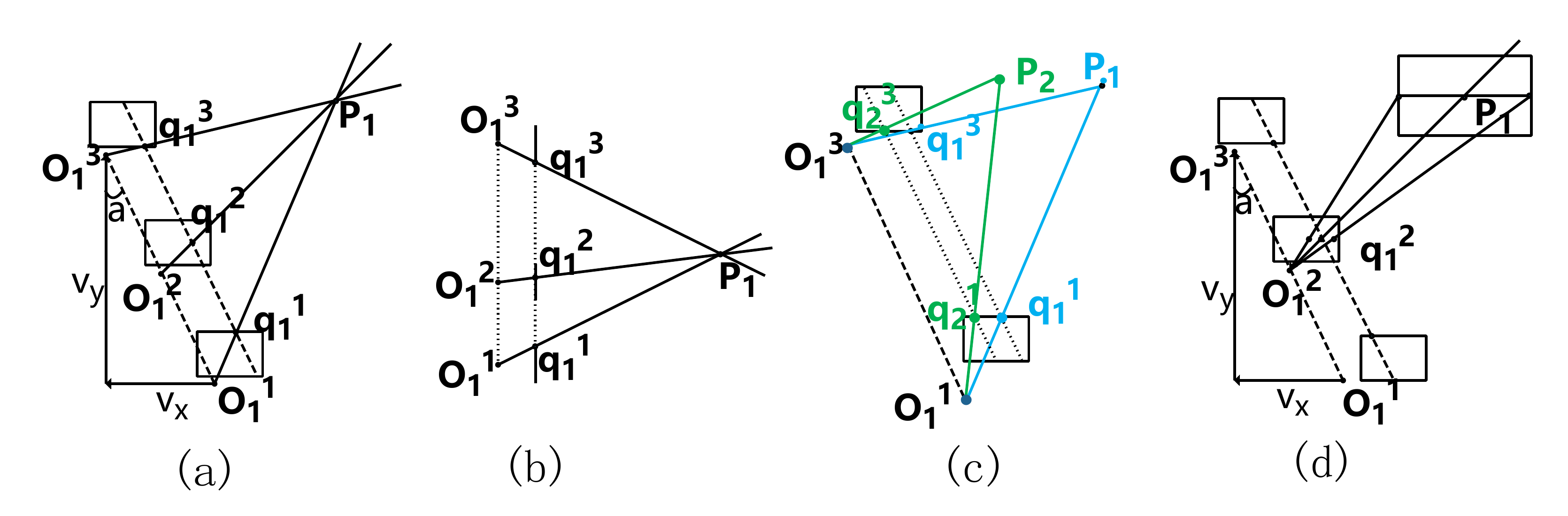}
\caption{In the case of X-Y pure translation, optimization compresses points into a straight line. In a single view, pixel points parallel to the motion trajectory back-project to the same 3D point ((a) and (b)). (c) Different intersection points in space are equidistant from the camera plane. (d) All intersection points formed by back-projection constitute a straight line.}
\label{fig}\label{fig:proof_degeneracy3}
\end{figure}
We present a pure translation scenario without the z-direction, suitable for multi-view BA, compressing space points into a straight line. Unlike the second type of degeneration, this does not depend on the number of images.

Firstly, considering a single view, as shown in Fig.\ref{fig:proof_degeneracy3}(1), during exposure, the camera moves from position $\mathbf{O}_{1}^{1}$ to $\mathbf{O}_{1}^{3}$, where pixel $\mathbf{q}_{1}^{1}$ is located on the first row and $\mathbf{q}_{1}^{3}$ is located on the $H$-th row. Their corresponding spatial points are located on the rays $\mathbf{O}_{1}^{1}\mathbf{q}_{1}^{1}$ and $\mathbf{O}_{1}^{3}\mathbf{q}_{1}^{3}$, intersecting at point $\mathbf{P}_{1}$. Actually, any point on the line $\mathbf{q}_{1}^{1}\mathbf{q}_{1}^{3}$ intersects with its corresponding 3D point at $\mathbf{P}_{1}$ during exposure. Let $\mathbf{O}_{1}^{2}$ be the camera position when exposed to the $h_{1}^{2}$-th row, with corresponding feature point $\mathbf{q}_{1}^{2}$. We only consider the plane $\mathbf{O}_{1}^{1}\mathbf{P}_{1}\mathbf{O}_{1}^{3}$ in Fig.\ref{fig:proof_degeneracy3}(2). The motion of $\mathbf{O}$ is uniform rectilinear motion thus, there exists a length relationship:
\begin{equation}
\left\{
\begin{array}{rcl}
l_{\mathbf{q}_1^1\mathbf{q}_1^2} = \frac{h_1^2}{H}l_{\mathbf{q}_1^1\mathbf{q}_1^3}, \quad l_{\mathbf{q}_3^1\mathbf{q}_1^2} = (1-\frac{h_1^2}{H})l_{\mathbf{q}_1^1\mathbf{q}_1^3} \\
l_{\mathbf{O}_1^1\mathbf{O}_1^2} = \frac{h_1^2}{H}l_{\mathbf{O}_1^1\mathbf{O}_1^3}, \quad l_{\mathbf{O}_3^1\mathbf{O}_1^2} = (1-\frac{h_1^2}{H})l_{\mathbf{O}_1^1\mathbf{O}_1^3}
\end{array} \right. 
\end{equation}
so we have:
\begin{equation}
\frac{l_{\mathbf{q}_{1}^{1}\mathbf{q}_{1}^{2}}}{l_{\mathbf{O}_{1}^{1}\mathbf{O}_{1}^{2}}} = \frac{l_{\mathbf{q}_{1}^{3}\mathbf{q}_{1}^{2}}}{l_{\mathbf{O}_{3}^{1}\mathbf{O}_{1}^{2}}} = \frac{l_{\mathbf{q}_{1}^{1}\mathbf{q}_{1}^{3}}}{l_{\mathbf{O}_{1}^{1}\mathbf{O}_{1}^{3}}}
\end{equation}

Therefore, the ray $\mathbf{O}_{1}^{2}\mathbf{q}_{1}^{2}$ will also intersect at point $\mathbf{P_{1}}$.
For every pixel on a line parallel to $\mathbf{O}_{1}^{1}\mathbf{O}_{1}^{3}$, it can be traced back to a different spatial point. For pixel points on different lines parallel to the motion trajectory, as shown in Fig.~\ref{fig:proof_degeneracy3}(3), because the distance $l_{q^1q^3}$ is the same and the camera focal length remains unchanged, according to similar triangles, these spatial points have equal distances to the image plane. Simultaneously, due to symmetry, these points also lie in the plane that is perpendicular to and bisects the line segment $\mathbf{O}_{1}^{1}\mathbf{O}_{1}^{2}$. Therefore, the intersection line of these two planes represents the final degenerated configuration, which is a line parallel to the x-axis, as shown in Fig.\ref{fig:proof_degeneracy3}(4).
\begin{figure}[t]
\centering
\includegraphics[width=0.7\columnwidth]{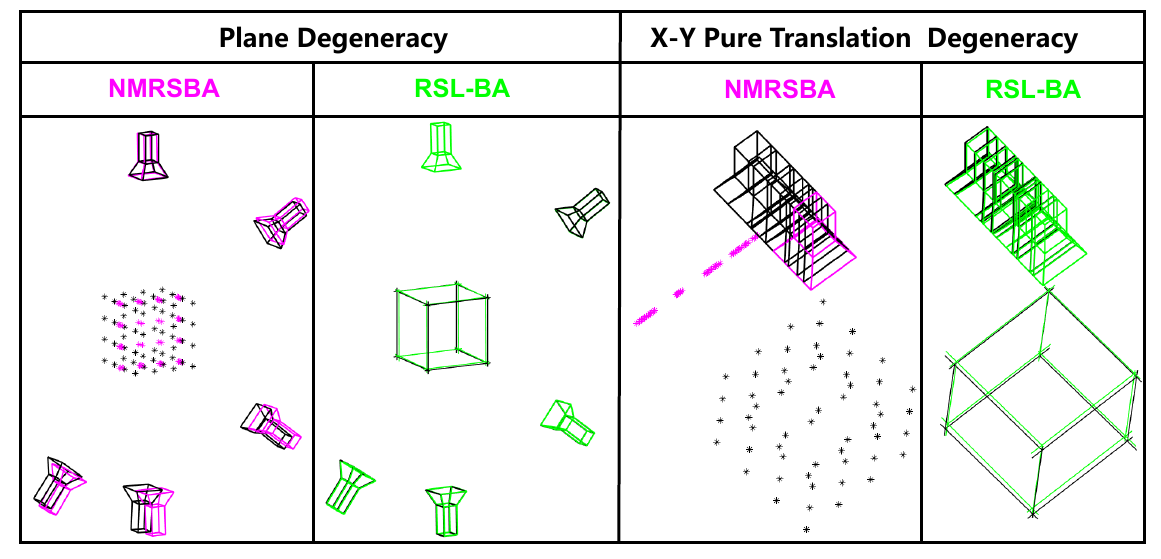}
\caption{
Comparison of degenerate resistance ability between NMRSBA~\cite{Albl2016} and proposed \textit{RSL-BA}. Ground truth camera poses and structure are colored with black. This example illustrates that our  proposed \textit{RSL-BA} has the resistance ability against the plane and X-Y pure translation degeneracy.
}
\label{fig: proof_degeneracy_experiment}
\end{figure}
In multiple views, with no z-direction movement, the imaging planes of different cameras coincide. We assign the same camera configuration to different views, including rotation, translation, angular velocity, and linear velocity. From the single-view case, corresponding points on the pixel plane parallel to the motion trajectory reconstruct to the same spatial point, with zero reprojection error.

For the line constraints proposed in this paper, \textit{RSL-BA} can still effectively suppress degeneration. The proof process is similar to the previous section Sec.~\ref{sec: 2-views Pure Translation}, with the only modification being setting $d_3=0$ in the camera's motion direction. As verified in Fig~\ref{fig: proof_degeneracy_experiment}, our proposed \textit{RSL-BA} can suppress X-Y pure translation degeneration, whereas conventional rolling shutter point-based bundle adjustment method NMRSBA~\cite{Albl2016} converges to a degenerate solution.

\begin{figure}[tb]
\centering
\includegraphics[width=0.9\columnwidth]{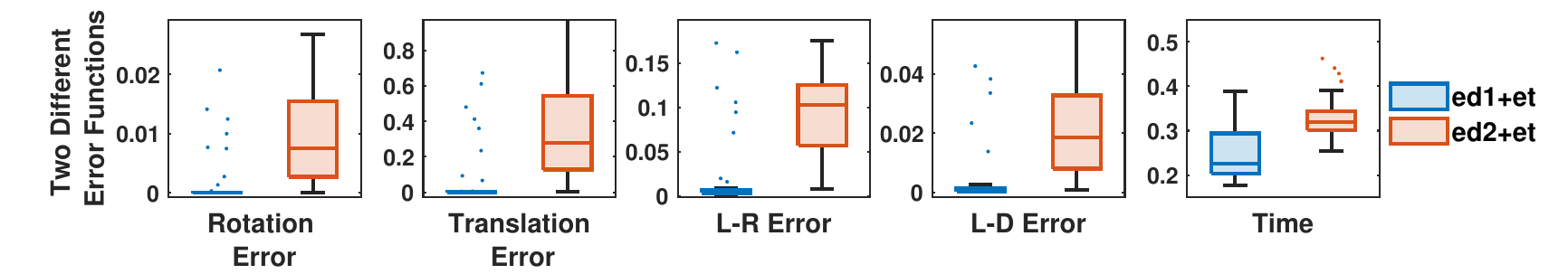}
\caption{Accuracy and computational time comparison of the proposed two combinations of reprojection errors.}
\label{fig: two_l_error}
\end{figure}
\section{Experiments}
\label{sec:Experiments}
We compare our method with two SOTA GS-based-method: \textit{1)} GSBA~\cite{lourakis2009sba}, \textit{2)} GLBA~\cite{taylor1995structure}, and four SOTA RS-based-method: \textit{1)} MRSBA~\cite{duchamp2015rolling}, \textit{2)} NMRSBA~\cite{Albl2016}, \textit{3)} WMRSBA\cite{NWMRSBA},  \textit{4)} NWRSBA\cite{NWMRSBA}. The experiments are conducted on a laptop with an Intel i7 CPU and all algorithms are implemented in MATLAB.
\subsubsection{Evaluation metrics.}
The accuracy measures employed encompass standard metrics such as rotation error, translation error, and algorithm efficiency. When conducting experiments related to line BA, we also incorporate error metrics that assess the accuracy of line reconstruction. Given the ground truth rotation $\mathbf{R}_{g}$, translation $\mathbf{t}_{g}$, and space line $\mathbf{L}_{g} = (\mathbf{n}^{\top}_{g}, \mathbf{a}^{\top}_{g})^{\top}$ with an arbitrary point $\mathbf{P}_{g}$ on it. Similarly, the optimized parameters are denoted as $\mathbf{R}_{d}$, $\mathbf{t}_{d}$,  $\mathbf{L}_{d} = (\mathbf{n}^{\top}_{d}, \mathbf{a}^{\top}_{d})^{\top}$, $\mathbf{P}_{d}$, respectively. The evaluation metrics are defined as:
\begin{equation}
\left\{
\begin{array}{rcl}
e_{Rotation} = arccos(\frac{Trace(\mathbf{R}_{g}^{\top}\mathbf{R}_{d})-1}{2})\\
e_{Translation} = 
arccos(\mathbf{t}_{g}^{\top}\mathbf{t}_{d}/||\mathbf{t}_{g}||\cdot||\mathbf{t}_{d}||)\\
e_{L-R} = 
arccos(\mathbf{a}_{g}^{\top}\mathbf{a}_{d}/||\mathbf{a}_{g}||\cdot||\mathbf{a}_{d}||)\\
e_{L-D} = \frac{\left|(\mathbf{a}_{g} \times \mathbf{a}_{d})( \mathbf{P}_{d} - \mathbf{P}_{g}  )\right|}{\| \mathbf{a}_{g} \times \mathbf{a}_{d} \|}
\end{array} \right. 
\end{equation}
\subsection{Synthetic Experiments}
\label{sec: Synthetic Experiments}
\subsubsection{Which error term in Sec.~\ref{subsec:Error_Definition} is better?}
In the synthetic environment, We set up a cubic box with 12 edges in 3D space and simulated a series of cameras around it to perform RS imaging. The experimental results are shown in Fig. \ref{fig: two_l_error}. It is evident that, the first type of reprojection error combination performs much better than the second combination in terms of accuracy and time. Therefore, in the forthcoming experiments, we shall exclusively employ the first type of reprojection error combination for comparison.

\begin{table}[t]
\centering
\caption{Robustness analysis against multiple levels of Gaussian noise (px).}
\resizebox{=0.7\textwidth}{!}{
\begin{tabular}{c  c  c  c  c  c }
\hline
noise level (px)    &  0.1 & 0.5 &  1.0 & 1.5 & 2.0 \\
\hline
Rotation Error    &  7.96e-6 & 9.24e-6  & 1.50e-5 & 1.71e-5 &  2.71e-5\\
\hline
Translation Error    &  2.32e-4 & 5.93e-4  & 8.91e-4 & 9.54e-4 &  1.55e-3\\
\hline
L-R Error    &  3.36e-3 & 3.83e-3  & 5.14e-3 & 5.22e-3 &  6.42e-3\\
\hline
L-D Error    &  2.22e-4 & 2.43e-4  & 3.90e-4 & 4.43e-4 &  4.75e-4\\
\hline
\end{tabular}
}
\label{table:robustness_experiments}
\end{table} 
\begin{figure}[t]
\centering
\includegraphics[width=.7\columnwidth]{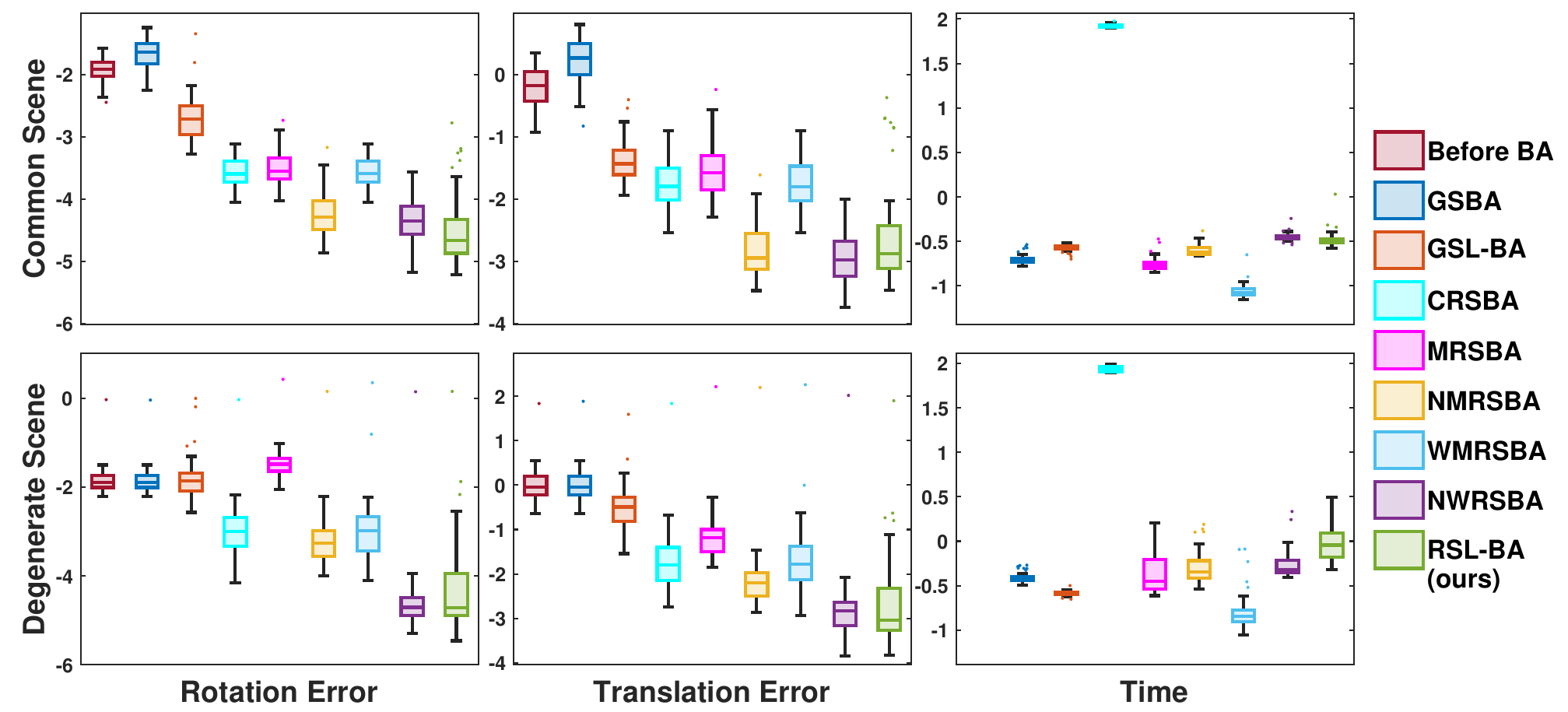}
\caption{
Comparison of accuracy and time among different algorithms.}
\label{fig: diff_function_ba}
\end{figure}
\subsubsection{How significant is the impact of noise on \textit{RSL-BA}?}

Following the same synthetic experimental setup as described above, we additionally introduce Gaussian noise to the endpoints of each line in this experiment. The noise levels are varied from 0.1 pixels to 2.0 pixels, and four different types of errors are recorded for the algorithm at each noise level. For each noise level, 50 experiments are conducted, and the median value is used as the final result. The results, presented in Table~\ref{table:robustness_experiments}, demonstrate that \textit{RSL-BA} provides stable estimations as the noise level increases from 0.1 pixels to 2 pixels.

\subsubsection{What are the advantages of \textit{RSL-BA} over point-based methods?}
Within this part, we will conduct a comparative analysis of our method alongside current SOTA methods, including point- and line-based methods. We set up a cubic box of points in the same positions as the 3D lines, totaling 56 points. To ensure fairness, we set up 8 lines with 7 points each to construct error functions, resulting in a total of precisely 56 points. Detailed experimental settings on the number of lines and the number of sampling points on each line are provided in the supplementary materials. We construct two scenarios: regular and classic scenarios with degeneration parallel to the y-axis. Due to significant differences in the experimental results of different methods, we uniformly take the logarithm (base 10) of the results for better observation. The experimental findings are shown in Fig. \ref{fig: diff_function_ba}. It is evident that in both situations, our \textit{RSL-BA} and the current NWRSBA exhibit comparable accuracy, outperforming alternative approaches based on points and lines. Additionally, the computation time is equivalent to that of NWRSBA. This indicates that the application of lines on Rolling Shutter camera holds significant promise.
\subsection{Synthetic Images}
\label{sec: Synthetic Images}
\begin{table*}[t]
\centering
\caption{The median absolute trajectory error (ATE) of different methods on WHU-RSVI~\cite{cao2020whu} dataset. The best and second results are shown in \textcolor{green}{green} and \textcolor{blue}{blue}, respectively.}
\resizebox{=.75\textwidth}{!}{
\begin{tabular}{c  c  c  c  c   }
\hline
    &  WHU-RSVI1 & WHU-RSVI2 &  WHU-RSVI3 & WHU-RSVI4 \\
\hline
GSBA    &  0.080992 & 0.061310  & \textcolor{blue}{0.030404} & 0.023698 \\
\hline
 GLBA   &  0.076173 & 0.065985 &  0.033554 & 0.024961 \\
\hline
NMRSBA & 0.050969 & \textcolor{blue}{0.041629}&  0.042317 & 0.028183 \\
\hline
NWRSBA   &  \textcolor{green}{0.040640}  & 0.045313 &  0.035451 & \textcolor{blue}{0.022666}\\
\hline
RSL-BA(ours)   &  \textcolor{blue}{0.0443502} & \textcolor{green}{0.039314 }&  \textcolor{green}{0.023351} & \textcolor{green}{0.020675} \\
\hline
\end{tabular}
}
\label{table:diff-method}
\end{table*}
In this section, we conduct experiments on input synthetic images. We use the WHU-RSVI~\cite{cao2020whu} dataset, from which we select two sets of data from trajectory1-fast and trajectory2-fast for 3D reconstruction and pose estimation. We first employ~\cite{purkait2017rolling} to detect RS curves by segmenting curves into multiple short-line segments and performing line fitting for initialization. The GS line-based SfM~\cite{liu20233d} is applied to initialize the \textit{RSL-BA} parameters. The comparative methods include GSBA, NMRSBA, NWRSBA, and GSLBA. Table~\ref{table:diff-method} shows the median absolute trajectory error of different methods, it can be observed that the proposed \textit{RSL-BA} method is the most stable one, achieving optimal or near-optimal results in all cases. Qualitative comparison are also provided in Fig.~\ref{fig: different_ba_result}. Unlike point-based methods, line-based methods often achieve good results with fewer feature lines. However, the GSL-BA method is not sufficiently stable when RS effects are prominent.

\subsection{Real Images}
\label{sec: Real Images}

\begin{figure}[tb]
\centering
\includegraphics[width=.7\columnwidth]{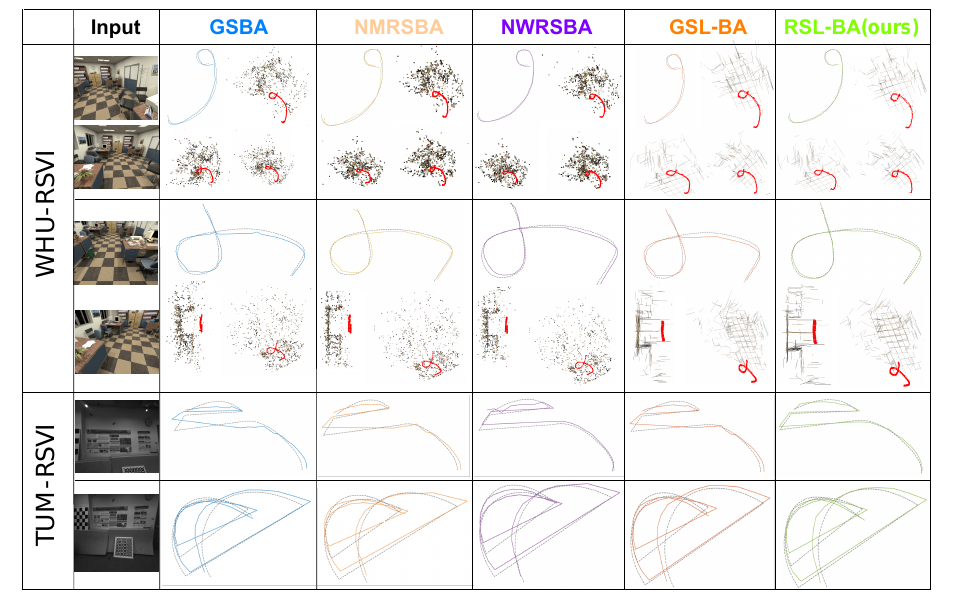}
\caption{Comparison of trajectories and 3D reconstruction on WHU-RSVI~\cite{cao2020whu} and TUM-RSVI\cite{schubert2019rolling} dataset. Each column represents a different bundle adjustment algorithm, and each row represents a different sequence. 
}
\label{fig: different_ba_result}
\end{figure}

In this section, we conduct experiments on the real image dataset TUM-RSVI~\cite{schubert2019rolling}. The experimental setup and comparison methods are similar to those in Sec.~\ref{sec: Synthetic Images}. Table~\ref{table:error_tum} and Fig~\ref{fig: different_ba_result} present some of the experimental results. As can be seen, in most cases, the \textit{RSL-BA} method outperforms other methods but is slightly weaker than NWRSBA. This is because the TUM-RSVI lacks line features and they are not visually prominent, making it less suitable for \textit{RSL-BA}.

\begin{table*}[t]
\centering
\caption{
The absolute trajectory error (ATE) comparison of different methods in TUM-RSVI\cite{schubert2019rolling} dataset. The best and second results are shown in \textcolor{green}{green} and \textcolor{blue}{blue}, respectively.
}
\resizebox{=0.65\textwidth}{!}{
\begin{tabular}{c  c  c  c  c  c c}
\hline
  & GSBA & GLBA &  NMRSBA & NWRSBA&  RSL-BA(ours) \\
\hline
 seq1 & 0.069484 & 0.086460 & 0.052366 & \textcolor{blue}{0.045064} & \textcolor{green}{0.037816} \\
\hline
 seq2 & 0.029821 & 0.030379 & 0.026028 & \textcolor{green}{0.023227} & \textcolor{blue}{0.025132} 
\\
 \hline
 seq3 & 0.065160 & 0.063718 & 0.057918 & \textcolor{blue}{0.055001} & \textcolor{green}{0.048187}  \\
 \hline
 seq4 & 0.049613 & 0.052026& 0.032214  & \textcolor{green}{0.030534} & \textcolor{blue}{0.031903}  \\
\hline
 seq5 & 0.031860 & 0.035839 & 0.019407 & \textcolor{green}{0.016066} & \textcolor{blue}{0.017659} \\
 \hline
 seq6 & 0.061966 & 0.061792 & 0.032434 & \textcolor{green}{0.024658} & \textcolor{blue}{0.026448} \\
 \hline
 seq7 & 0.051621 & 0.056534 & \textcolor{green}{0.039154} & \textcolor{blue}{0.039620} & 0.039701 \\
 \hline
 seq8 & 0.026403 & 0.028690 & \textcolor{green}{0.024807} & 0.025926 & \textcolor{blue}{0.024983}\\
 \hline
 seq9 & 0.098334 & 0.098212 & 0.082481 & \textcolor{green}{0.073580}  & \textcolor{blue}{0.080523}  \\
 \hline
 seq10 & 0.81174  & 0.81180 & 0.59390 & \textcolor{green}{0.53296} & \textcolor{blue}{0.57477} \\
\hline
\end{tabular}
}
\label{table:error_tum}
\end{table*} 

\section{Conclusion}
\label{sec:Conclusion}
This paper presents the first solution of line-based \textit{RSL-BA}. By utilizing points and tangent directions on curves, we have established a series of faster and more robust curve-to-line re-projection errors. The proposed \textit{RSL-BA} method can prevent three common degeneracies in RSBA, one of which is newly introduced by us. Extensive experiments in real and synthetic data verify the effectiveness and efficiency of the proposed \textit{RSL-BA} method.

\section*{Acknowledgments}
This work is supported by the National Key R$\&$D Program of China (No. 2022ZD01190030), Nature Science Foundation of China (No. 62102145), Jiangxi Provincial 03 Special Foundation and 5G Program (Grant No. 20224ABC03A05), Lushan Lab Research Funding, and Changsha Science Fund for Distinguished Young Scholars (kq2306002). 

%
%
\bibliographystyle{splncs04}
\bibliography{main}
\end{document}


\title{Supplementary Material \\ RSL-BA: Rolling Shutter Line Bundle Adjustment} 

\titlerunning{RSL-BA}

\author{Yongcong Zhang\inst{1,3,*}\orcidlink{0009-0009-1254-6549} \and 
Bangyan Liao\inst{2,*}\orcidlink{0009-0007-7739-4879} \and
Yifei Xue\inst{1,3}\orcidlink{0000-0002-4443-4367} \and
Chen Lu\inst{4}\orcidlink{0009-0003-5779-4673} \and \\
Peidong Liu\inst{2}\orcidlink{0000-0002-9767-6220} \and
Yizhen Lao\inst{1,\dagger}\orcidlink{0000-0002-6284-1724}
}   

\authorrunning{Y. Zhang, B. Liao et al.}

\institute{\textsuperscript{1}Hunan University \quad \textsuperscript{2}Westlake University \quad \textsuperscript{3}DaHe.AI \\ \textsuperscript{4}Dreame Technology Company}
\maketitle
\renewcommand{\thefootnote}{\relax}\footnotetext{\textsuperscript{$\star$} Equal contribution\\  
  \inst{\dagger} Corresponding author: (yizhenlao@hnu.edu.cn) \\
  Project page: \href{https://github.com/zhangtaxue/RSL-BA}{https://github.com/zhangtaxue/RSL-BA}}
\renewcommand{\thefootnote}{\arabic{footnote}}
\setcounter{footnote}{0}

\section*{Overview}
In this supplementary material, we further discuss the following content:
\begin{itemize}
    \item The transformation between Plücker coordinates and orthogonal representation(Sec.~\ref{supp:sec: p2o}).
    \item Jacobian derivation for RSL-BA(Sec.~\ref{supp:sec:Jacobian}).
    \item The impact of the number of feature lines and points(Sec.~\ref{supp:sec: Synthetic Experiments}).
    \\ $-$ What is the optimal number of points to measure along a line?(Sec.~\ref{supp:sec: number of points}).
    \\ $-$ What is the optimal number of lines to employ for RSL-BA?(Sec.~\ref{supp:sec:number of lines}).
    \item Complete results on the TUM-RSVI and WHU-RSVI dataset(Sec.~\ref{supp:sec: Complete results}).
    \\ $-$ Synthetic Images.(Sec.~\ref{supp: sec: Synthetic Images}).
    \\ $-$ Real Images.(Sec.~\ref{supp:sec:Real Images}).
\end{itemize}

\section{The Transformation between Plücker Coordinates and Orthogonal Representation}
\label{supp:sec: p2o}
The Plücker coordinate of the line are defined as: $\mathbf{L} = (\mathbf{n}^{\top}, \mathbf{a}^{\top})^{\top}$ with the orthogonal representation parameters $\boldsymbol{\tau} = [\psi_1,\psi_2, \psi_3,\phi]^{\top}$. Where $\mathbf{a} \in \mathbf{R}^3$ represents the direction vector of the line, $\mathbf{n} \in \mathbf{R}^3$ represents the normal vector. We have\cite{he2018pl,bartoli2005structure}: 
\begin{equation}\label{equ:p2o1}
\mathbf{U} = \mathbf{Exp}([\psi_1,\psi_2, \psi_3]^{\land}) = \begin{bmatrix} \frac{\mathbf{n}}{\left\lVert \mathbf{n} \right\rVert} \quad \frac{\mathbf{a}}{\left\lVert \mathbf{a} \right\rVert} \quad \frac{\mathbf{n \times a}}{\left\lVert \mathbf{n \times a} \right\rVert}
\end{bmatrix}
\end{equation}

The function $\mathbf{Exp}$ maps from $\mathfrak{so}(3)$ to $\mathbf{SO}(3)$, and $\boldsymbol{\psi} = [\psi_1, \psi_2, \psi_3]^{\top}$ represents the rotation angles from the camera coordinate system to the line coordinate system around the x, y, and z axes, respectively. By utilizing equation Eq.~(\ref{equ:p2o1}), we can obtain the first term of the orthogonal representation. 

\begin{equation}\label{equ:p2o2}
\mathbf{W} = 
\begin{bmatrix}
w_1 & -w_2 \\
w_2 & w_1
\end{bmatrix} = 
\begin{bmatrix} cos(\phi) & -sin(\phi) \\
                sin(\phi) & cos(\phi)
\end{bmatrix}
=
\frac{1}{\sqrt{\left\lVert \mathbf{n} \right\rVert^2+\left\lVert \mathbf{a} \right\rVert^2}}\begin{bmatrix}
\left\lVert \mathbf{n} \right\rVert & -\left\lVert \mathbf{a} \right\rVert \\
\left\lVert \mathbf{a} \right\rVert & \left\lVert \mathbf{n} \right\rVert
\end{bmatrix}
\end{equation}
With Eq.~(\ref{equ:p2o2}), we can obtain the second term of the orthogonal representation.

The transformation from the orthogonal representation to Plücker coordinates can be computed as follows:
\begin{equation}
\mathbf{L'} = [w_1\mathbf{u}_1^{\top}, w_2\mathbf{u}_2^{\top}]^{\top} = \frac{1}{\sqrt{\left\lVert \mathbf{n} \right\rVert^2+\left\lVert \mathbf{a} \right\rVert^2}} \mathbf{L}
\end{equation}
$\mathbf{L}'$ and $\mathbf{L}$ differ by a scale factor, but represent the same line.

\section{Jacobian Derivation for RSL-BA}
\label{supp:sec:Jacobian}
Since lines in space only have four degrees of freedom, and Plücker coordinates are over-parameterized, they cannot be directly used for unconstrained optimization. Therefore, we often use Plücker coordinates for initialization and transformation, while employing an orthogonal representation for parameter optimization. 
Let the representation of the space line $\mathbf{L}$ in the world coordinate system be $\mathbf{L}_w = (\mathbf{n}^{\top}_w, \mathbf{a}^{\top}_w)^{\top}$, where $\mathbf{n}_w$ and $\mathbf{a}_w$ respectively denote the normal vector to the line and the direction of the line from the camera center. The representation of the line $\mathbf{L}$ in the camera coordinate system is $\mathbf{L}_c = (\mathbf{n}^{\top}_c, \mathbf{a}^{\top}_c)^{\top}$. The matrix from the world coordinate system to the camera when the camera exposes the $v$-th row of pixels is:

\begin{equation}
\begin{split}
\mathbf{T}_{cw}^{v} = \begin{bmatrix}
(\mathbf{I}+v[\boldsymbol{\omega}]_{\times})\mathbf{R}_{cw} \quad & \mathbf{t}_{cw}+v\mathbf{d} \\
0 \quad & 1
\end{bmatrix}
\end{split}
\end{equation}

The transformation of Plücker line coordinates from the world coordinate system to the camera coordinate system is denoted as $\mathbf{N}_{cw}^{v}$:
\begin{equation}
\begin{split}
\mathbf{N}_{cw}^{v} = \begin{bmatrix}
(\mathbf{I}+v[\boldsymbol{\omega}]_{\times})\mathbf{R}_{cw} \quad & [\mathbf{t}_{cw}+v\mathbf{d}]_{\times}(\mathbf{I}+v[\boldsymbol{\omega}]_{\times})\mathbf{R}_{cw} \\
\mathbf{0} \quad & (\mathbf{I}+v[\boldsymbol{\omega}]_{\times})\mathbf{R}_{cw}
\end{bmatrix}
\end{split}
\end{equation}

We have:
\begin{equation}
\begin{split}
\mathbf{L}_{c}^{v} = 
\mathbf{N}_{cw}^{v}\mathbf{L}_{w} =
\begin{bmatrix}
(\mathbf{I}+v[\boldsymbol{\omega}]_{\times})\mathbf{R}_{cw} \quad & [\mathbf{t}_{cw}+v\mathbf{d}]_{\times}(\mathbf{I}+v[\boldsymbol{\omega}]_{\times})\mathbf{R}_{cw} \\
\mathbf{0} \quad & (\mathbf{I}+v[\boldsymbol{\omega}]_{\times})\mathbf{R}_{cw}
\end{bmatrix}L_{w}
\end{split}
\end{equation}

Let the parametric equation of the curve under the aforementioned parameters be:
\begin{equation}
\begin{split}
l_{1}v^{3}+l_{2}uv^{2}+(l_{3}+l_{4})v^{2}+l_{5}uv+(l_{6}+l_{7})v+l_{8}u+l_{9}=0
\end{split}
\end{equation}

Take a point $q=[u \quad v]^{\top}$ from the projected curve.
Next, we will compute the Jacobian matrices of various error functions. Firstly, let's consider the perpendicular-distance error:
\begin{equation}
\begin{split}
\mathbf{e}_{d1} = \frac{|l_{1}v^{3}+l_{2}uv^{2}+(l_{3}+l_{4})v^{2}+l_{5}uv+(l_{6}+l_{7})v+l_{8}u+l_{9}|}{\sqrt{(l_{8}+vl_{5}+v^2l_{2})^{2}+(l_{6}+vl_{3}+v^2l_{1})^{2}}}
\end{split}
\end{equation}

According to the chain rule for differentiation, the Jacobian matrix is represented as\cite{he2018pl,bartoli2005structure}:
\begin{equation}
\begin{split}
\mathbf{J}_{\mathbf{e}_{d1} }=\frac{\partial \mathbf{e}_{d1} }{\partial \mathbf{s_c}}\frac{\partial \mathbf{s_c}}{\mathbf{L}_{c}^{v}}[\frac{\partial \mathbf{L}_{c}^{v}}{\partial \delta \mathbf{x}} \quad \frac{\partial \mathbf{L}_{c}^{v}}{\partial \mathbf{L}_{w}}\frac{\partial \mathbf{L}_{w}}{\partial \delta \boldsymbol{\tau}}]
\end{split}
\end{equation}

The first term represents the partial derivative of the error with respect to the curve parameters, while the second term represents the partial derivative of the curve parameters with respect to the line features in the camera coordinate system. 
The last term in the matrix contains two parts: one is the derivative of the rotation, translation, angular velocity and linear velocity with respect to the line features in the camera coordinate system, and the other is the derivative of the four parameters increment with respect to the line in its orthogonal representation. 

The first term:
\begin{equation}\label{equ: J_1}
\begin{split}
\frac{\partial \mathbf{e}_{d1} }{\partial \mathbf{s_{c}}}=[\frac{\partial \mathbf{e}_{d1} }{\partial l_1} \quad \frac{\partial \mathbf{e}_{d1} }{\partial l_2} \quad \frac{\partial \mathbf{e}_{d1} }{\partial l_3} \quad \frac{\partial \mathbf{e}_{d1} }{\partial l_4} \quad \frac{\partial \mathbf{e}_{d1} }{\partial l_5} \quad \frac{\partial \mathbf{e}_{d1} }{\partial l_6} \quad \frac{\partial \mathbf{e}_{d1} }{\partial l_7} \quad \frac{\partial \mathbf{e}_{d1} }{\partial l_8} \quad \frac{\partial \mathbf{e}_{d1} }{\partial l_9}]_{1 \times 9}
\end{split}
\end{equation}

The second term:
\begin{equation}
\begin{split}
\frac{\partial \mathbf{s_{c}}}{\partial \mathbf{L}_{c}^{v}}=[\frac{\partial \mathbf{s_{c}}}{\partial \mathbf{n}_c} \quad \frac{\partial \mathbf{s_{c}}}{\partial \mathbf{a}_c}] 
= \begin{bmatrix}
\frac{\partial l_1}{\partial n_{1}} & \frac{\partial l_1}{\partial n_{2}} &...\frac{\partial l_1}{\partial a_{3}} \\
\frac{\partial l_2}{\partial n_{1}} & \frac{\partial l_2}{\partial n_{2}} &... \frac{\partial l_2}{\partial a_{3}} \\
......\\
\frac{\partial l_9}{\partial n_{1}} & \frac{\partial l_9}{\partial n_{2}} &... \frac{\partial l_9}{\partial a_{3}} \\
\end{bmatrix}_{9 \times 6}
\end{split}
\end{equation}

The first term in the parentheses:
\begin{equation}
\begin{split}
\mathbf{L}_{c}^{v} = 
\begin{bmatrix}
(\mathbf{I}+v[\boldsymbol{\omega}]_{\times})\mathbf{R}_{cw} & [\mathbf{t}_{cw}+v\mathbf{d}]_{\times}(\mathbf{I}+v[\boldsymbol{\omega}]_{\times})\mathbf{R}_{cw} \\
\mathbf{0} & (\mathbf{I}+v[\boldsymbol{\omega}]_{\times})\mathbf{R}_{cw}
\end{bmatrix}L_{w} 
\\ =
\begin{bmatrix}
(\mathbf{I}+v[\boldsymbol{\omega}]_{\times})\mathbf{R}_{cw}\mathbf{n}_{w} + [\mathbf{t}_{cw}+v\mathbf{d}]_{\times}(\mathbf{I}+v[\boldsymbol{\omega}]_{\times})\mathbf{R}_{cw}\mathbf{a}_{w} \\
 (\mathbf{I}+v[\boldsymbol{\omega}]_{\times})\mathbf{R}_{cw}\mathbf{a}_{w}
\end{bmatrix}
\end{split}
\end{equation}

We first differentiate with respect to rotation:
\begin{equation}
\begin{split}
\frac{\partial \mathbf{L}_{c}^{v}}{\partial \delta \boldsymbol{\theta}} =\begin{bmatrix}
\frac{(\mathbf{I}+v[\boldsymbol{\omega}]_{\times})(\mathbf{I}+[\delta \boldsymbol{\theta}]_{\times})\mathbf{R}_{cw}n_{w} + [\mathbf{t}_{cw}+v\mathbf{d}]_{\times}(\mathbf{I}+v[\boldsymbol{\omega}]_{\times})(\mathbf{I}+[\delta \boldsymbol{\theta}]_{\times})\mathbf{R}_{cw}\mathbf{a}_{w}}{\partial \delta \boldsymbol{\theta}} \\
 \frac{(\mathbf{I}+v[\boldsymbol{\omega}]_{\times})(\mathbf{I}+[\delta \boldsymbol{\theta}]_{\times})\mathbf{R}_{cw}\mathbf{a}_{w}}{\partial \delta \boldsymbol{\theta}}
\end{bmatrix}
\end{split}
\end{equation}

We observe that all of them have the form $\frac{\partial(A[\delta \boldsymbol{\theta}]_{\times}b)}{\partial \delta \boldsymbol{\theta}}$, where $A$ is a $3 \times 3$ matrix, and $b$ is a $3 \times 1$ matrix:
\begin{equation}
\begin{split}
\mathbf{A} =\begin{bmatrix}
A_1 & A_2 & A_3 \\
A_4 & A_5 & A_6 \\
A_7 & A_8 & A_9
\end{bmatrix}
,
[\delta \boldsymbol{\theta}]_{\times} =
\begin{bmatrix}
0 & -\delta \theta_3 & \delta \theta_2 \\
\delta \theta_3 & 0 & -\delta \theta_1 \\
-\delta \theta_2 & \delta \theta_1 & 0
\end{bmatrix}
,
\mathbf{b} =\begin{bmatrix}
b1 \\
b2 \\
b3
\end{bmatrix}
\end{split}
\end{equation}

Expanding $\frac{\partial(\mathbf{A}[\delta \boldsymbol{\theta}]_{\times}\mathbf{b})}{\partial \delta \boldsymbol{\theta}}$, we get:
\begin{equation}
\begin{split}
\frac{\partial(\mathbf{A}[\delta \boldsymbol{\theta}]_{\times}\mathbf{b})}{\partial \delta \boldsymbol{\theta}} =
\begin{bmatrix}
(A_3b_2 - A_2b_3) & (A_1b_3 - A_3b_1) & (A_2b_1 - A_1b_2)\\
(A_6b_2 - A_5b_3) & (A_4b_3 - A_6b_1) & (A_5b_1 - A_4b_2)\\
(A_9b_2 - A_8b_3) & (A_7b_3 - A_9b_1) & (A_8b_1 - A_7b_2)
\end{bmatrix}
\end{split}
\end{equation}

Differentiate with respect to translation:
\begin{equation}
\begin{split}
\frac{\partial \mathbf{L}_{c}^{v}}{\partial \mathbf{t}_{cw}} =
\begin{bmatrix}
\frac{\partial((\mathbf{I}+v[\boldsymbol{\omega}]_{\times})\mathbf{R}_{cw}\mathbf{n}_{w} + [\mathbf{t}_{cw}+v\mathbf{d}]_{\times}(\mathbf{I}+v[\boldsymbol{\omega}]_{\times})\mathbf{R}_{cw}\mathbf{a}_{w})}{\partial \mathbf{t}_{cw}} \\
 \frac{\partial( (\mathbf{I}+v[\boldsymbol{\omega}]_{\times})\mathbf{R}_{cw}\mathbf{a}_{w})}{\partial \mathbf{t}_{cw}}
\end{bmatrix}
\\
=
\begin{bmatrix}
\frac{\partial( [\mathbf{t}_{cw}]_{\times}(\mathbf{I}+v[\boldsymbol{\omega}]_{\times})\mathbf{R}_{cw}\mathbf{a}_{w})}{\partial \mathbf{t}_{cw}} \\
\mathbf{0}
\end{bmatrix}
=
-\begin{bmatrix}
[(\mathbf{I}+v[\boldsymbol{\omega}]_{\times})\mathbf{R}_{cw}a_{w}]_{\times} \\
\mathbf{0}
\end{bmatrix}_{6 \times 3}
\end{split}
\end{equation}

Differentiate with respect to angular velocity:
\begin{equation}
\begin{split}
\frac{\partial \mathbf{L}_{c}^{v}}{\partial \boldsymbol{\omega}} =
\begin{bmatrix}
\frac{\partial((\mathbf{I}+v[\boldsymbol{\omega}]_{\times})\mathbf{R}_{cw}\mathbf{n}_{w} + [\mathbf{t}_{cw}+v\mathbf{d}]_{\times}(\mathbf{I}+v[\boldsymbol{\omega}]_{\times})\mathbf{R}_{cw}\mathbf{a}_{w})}{\partial \boldsymbol{\omega}} \\
 \frac{\partial( (\mathbf{I}+v[\boldsymbol{\omega}]_{\times})\mathbf{R}_{cw}\mathbf{a}_{w})}{\partial \boldsymbol{\omega}}
\end{bmatrix}
\end{split}
\end{equation}

There are two forms: $[\boldsymbol{\omega}]{\times}\mathbf{b}$ and $\mathbf{A}[\boldsymbol{\omega}]{\times}\mathbf{b}$. The forms in the differentiation with respect to rotation and translation are the same. Here we will not expand it in detail.

Differentiate with respect to linear velocity:
\begin{equation}
\begin{split}
\frac{\partial \mathbf{L}_{c}^{v}}{\partial \mathbf{d}} =
\begin{bmatrix}
\frac{\partial((\mathbf{I}+v[\boldsymbol{\omega}]_{\times})\mathbf{R}_{cw}\mathbf{n}_{w} + [\mathbf{t}_{cw}+v\mathbf{d}]_{\times}(\mathbf{I}+v[\boldsymbol{\omega}]_{\times})\mathbf{R}_{cw}\mathbf{a}_{w})}{\partial \mathbf{d}} \\
 \frac{\partial( (\mathbf{I}+v[\boldsymbol{\omega}]_{\times})\mathbf{R}_{cw}\mathbf{a}_{w})}{\partial \mathbf{d}}
\end{bmatrix}
\\=
\begin{bmatrix}
\frac{\partial( v[\mathbf{d}]_{\times}(\mathbf{I}+v[\boldsymbol{\omega}]_{\times})\mathbf{R}_{cw}\mathbf{a}_{w})}{\partial \mathbf{d}} \\
\mathbf{0}
\end{bmatrix}
=
\begin{bmatrix}
[v(\mathbf{I}+v[\boldsymbol{\omega}]_{\times})\mathbf{R}_{cw}\mathbf{a}_{w}]_{\times} \\
\mathbf{0}
\end{bmatrix}_{6 \times 3}
\end{split}
\end{equation}

The second term in the parentheses: 

\begin{equation}
\begin{split}
\mathbf{L}_{c}^{v} = 
\begin{bmatrix}
\mathbf{n}_{c} \\ \mathbf{a}_{c}
\end{bmatrix}
=
\mathbf{N}_{cw}^{v}\mathbf{L}_{w} =
\begin{bmatrix}
(\mathbf{I}+v[\omega]_{\times})\mathbf{R}_{cw} & [\mathbf{t}_{cw}+v\mathbf{d}]_{\times}(\mathbf{I}+v[\omega]_{\times})\mathbf{R}_{cw} \\
0 & (\mathbf{I}+v[\boldsymbol{\omega}]_{\times})\mathbf{R}_{cw}
\end{bmatrix}L_{w}
\end{split}
\end{equation}

So we have:
\begin{equation}
\frac{\partial \mathbf{L}_{c}^{v}}{\partial \mathbf{L}_{w}} = \mathbf{N}_{cw}^{v}
\end{equation}

The last term: 
\begin{equation}
\begin{split}
\frac{\partial \mathbf{L}_{w}}{\partial \delta \boldsymbol{\tau}} = 
\begin{bmatrix}
\frac{\partial \mathbf{L}_{w}}{\partial \psi_1} & \frac{\partial \mathbf{L}_{w}}{\partial \psi_2} & \frac{\partial \mathbf{L}_{w}}{\partial \psi_3} & \frac{\partial \mathbf{L}_{w}}{\partial \phi}
\end{bmatrix}
\\=
\begin{bmatrix}
0 & -W_1\mathbf{U}_3 & W_1\mathbf{U}_2 & -W_2\mathbf{U}_1 \\
W_2\mathbf{U}_3 & 0 & -W_2\mathbf{U}_1 & W_1\mathbf{U}_2
\end{bmatrix}_{6 \times 4}
\end{split}
\end{equation}

The derivation of the Jacobian matrices for the remaining error functions is similar to this one, except for the differentiation of the first term Eq.~(\ref{equ: J_1}) with respect to the curve parameters.

\section{The impact of the number of feature lines and points.}
\label{supp:sec: Synthetic Experiments}

In Sec~\ref{supp:sec: number of points}, we explore the impact of the number of sampling points on the curve on the optimization results. In Sec~\ref{supp:sec:number of lines}, we verify the effect of the number of sampling lines on the error.

\subsection{What is the optimal number of points to measure along a line?}
\label{supp:sec: number of points}

\begin{figure}[tb]
\centering
\includegraphics[width=1\columnwidth]{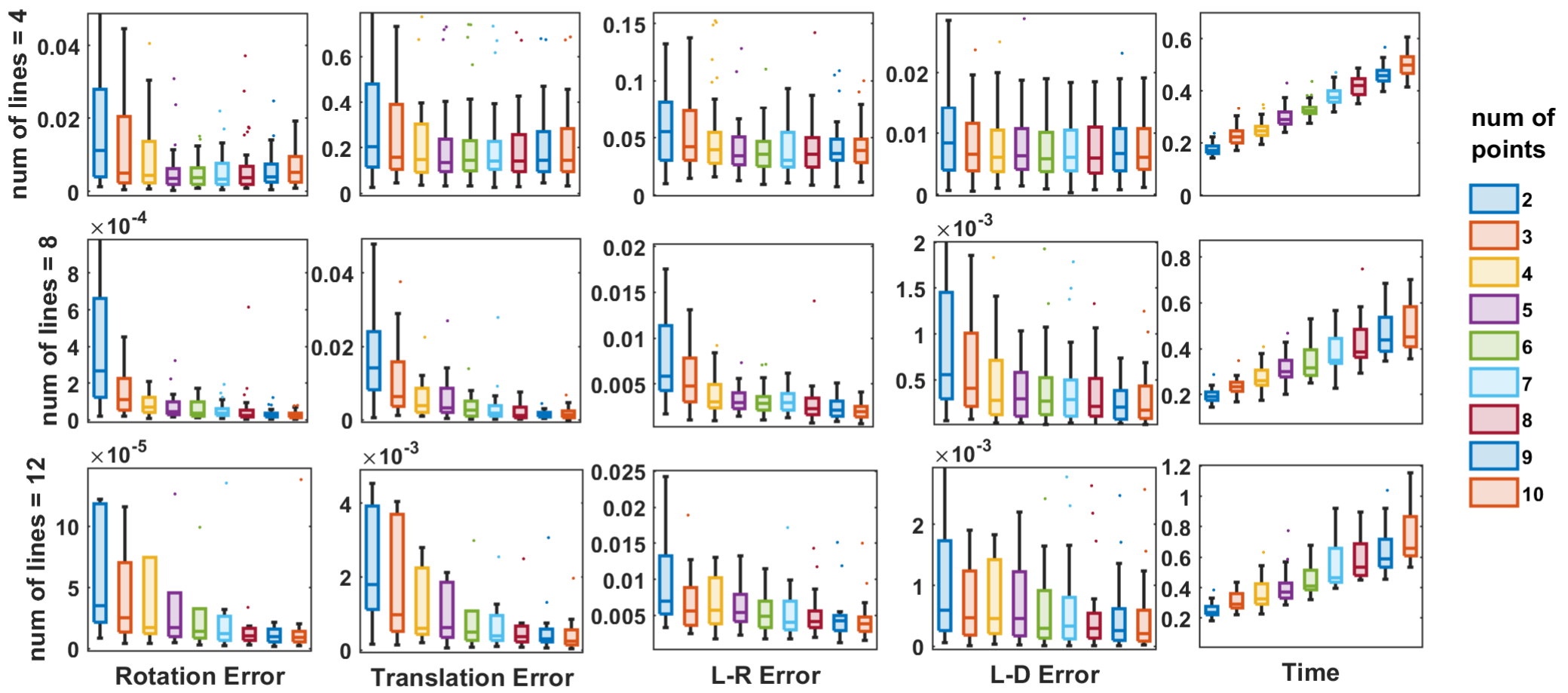}
\caption{The impact of the number of points on the curve on the accuracy and time of bundle adjustment.}
\label{fig: diff_l}
\end{figure}

In this section, we investigate the influence of the number of points taken on the curves on the accuracy of our algorithm.
Firstly, we establish a predetermined number of lines in space and then conduct experiments by sequentially taking 2 to 10 points on the curve. We iterate this procedure 50 times and draw box plots of the empirical outcomes using varying quantities of points. We perform three sets of tests using constant numbers of lines in space: 4, 8, and 12. The results are displayed in Fig. \ref{fig: diff_l}. As the number of points sampled on the lines rises, the precision of the experimental results progressively enhances, albeit at the cost of increased time consumption. However, after the number of points on the lines reaches approximately 5, the enhancement in accuracy becomes less notable while the time consumption steadily increases. The reason for this is that the curve expression has only 9 unknowns, and each point can impose two constraints: slope and distance. Therefore, the line constraints can be maximally effective when there are about 5 points on the line. Hence, we recommend using 4 to 6 points on the curves as constraints.
\begin{figure}[h]
\centering
\includegraphics[width=1\columnwidth]{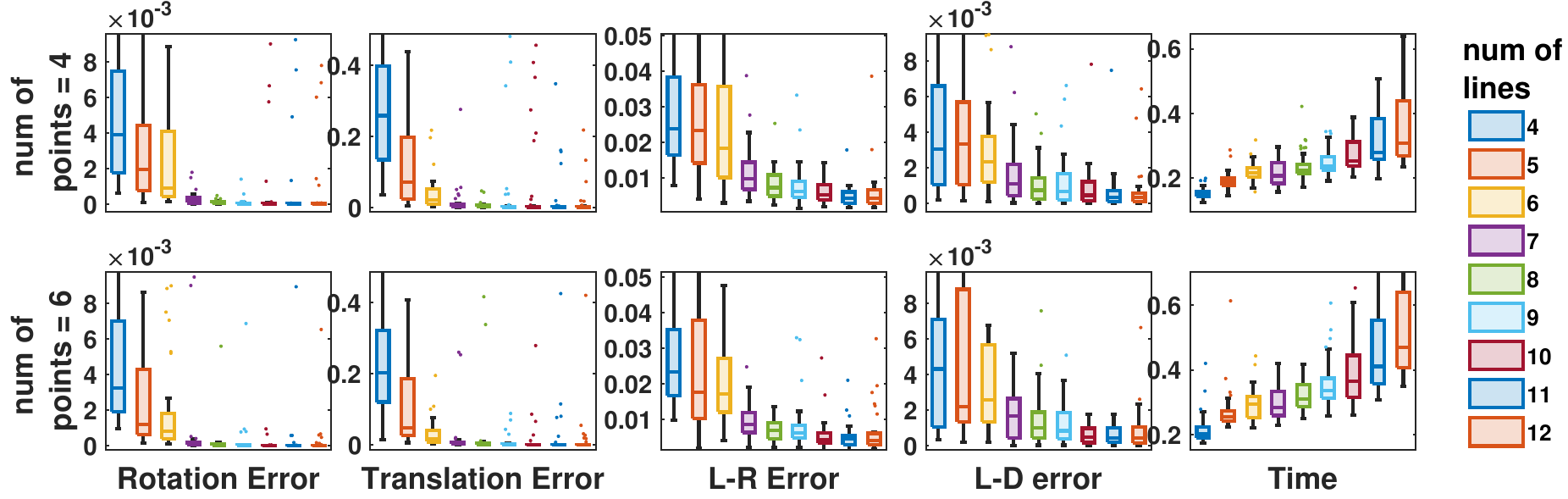}
\caption{
The impact of the number of 3D lines in space on the accuracy and time of bundle adjustment.}
\label{fig: diff_p}
\end{figure}

\subsection{What is the optimal number of lines to employ for RSL-BA?}
\label{supp:sec:number of lines}

Within this section, we shall examine the impact of the quantity of lines in 3D space on the accuracy of our algorithm.
We fix the number of points taken on each curve and subsequently conduct experiments by sequentially arranging 4 to 12 lines in space. We iterate this process 50 times and draw box plots of the experimental outcomes using varying quantities of lines. We conduct two sets of experiments with a fixed number of points on the curves: 4 points and 6 points. The results are displayed in Fig. \ref{fig: diff_p}. An increase in the number of lines leads to a noticeable enhancement in the algorithm's accuracy, followed by a period of stabilization. Meanwhile, the time consumption consistently rises.

\section{Complete results on the TUM-RSVI\cite{schubert2019rolling} and WHU-RSVI~\cite{cao2020whu} dataset}
\label{supp:sec: Complete results}

We compare our method with two SOTA GS-based-method: \textit{1)} GSBA~\cite{lourakis2009sba}, \textit{2)} GLBA~\cite{taylor1995structure}, and two SOTA RS-based-method: \textit{1)} NMRSBA~\cite{Albl2016}, \textit{2)} NWRSBA\cite{NWMRSBA}. The experiments are conducted on a laptop with an Intel i7 CPU and all algorithms are implemented in MATLAB.

\subsection{Synthetic Images}
\label{supp: sec: Synthetic Images}

\begin{figure}[tb]
\centering
\includegraphics[width=1\columnwidth]{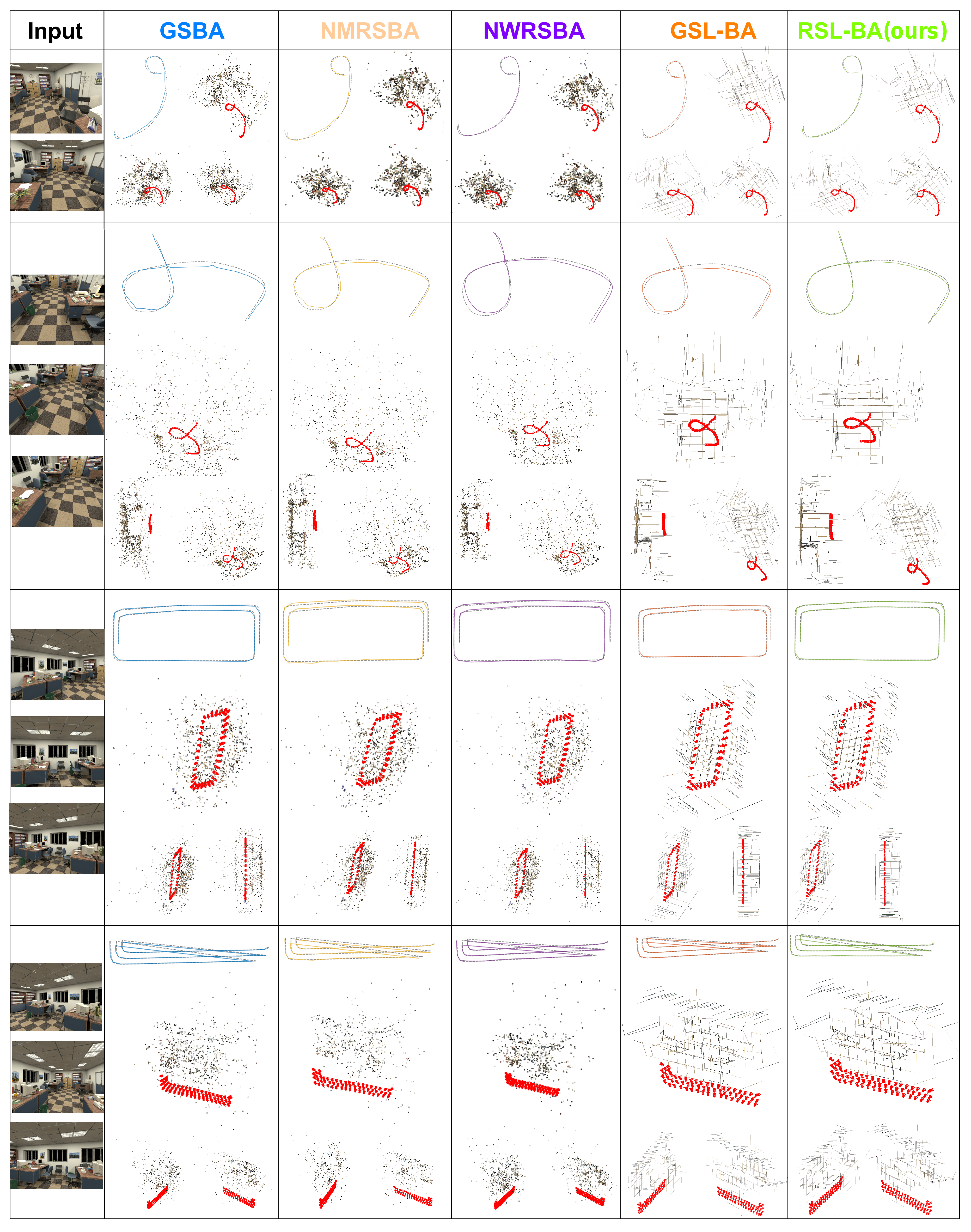}
\caption{Trajectories and 3D reconstruction comparison. Each column represents a different bundle adjustment algorithm, and each row represents a different sequence. It can be observed that our algorithm has smaller trajectory errors and better reconstruction results.}
\label{fig: different_ba_result_whu}
\end{figure}

\begin{table*}[t]
\centering
\caption{The median absolute trajectory error (ATE) of different methods on WHU-RSVI~\cite{cao2020whu} dataset, the best results are highlighted by bold font.}
\resizebox{=.7\textwidth}{!}{
\begin{tabular}{c  c  c  c  c   }
\hline
    &  WHU-RSVI1 & WHU-RSVI2 &  WHU-RSVI3 & WHU-RSVI4 \\
\hline
GSBA    &  0.080992 & 0.061310  & 0.030404 & 0.023698 \\
\hline
 GLBA   &  0.076173 & 0.065985 &  0.033554 & 0.024961 \\
\hline
NMRSBA & 0.050969 & 0.041629&  0.042317 & 0.028183 \\
\hline
NWRSBA   &  $\mathbf{0.040640}$  & 0.045313 &  0.035451 & 0.022666\\
\hline
RSL-BA(ours)   &  0.0443502 & $\mathbf{0.039314 }$&  $\mathbf{0.023351}$ & $\mathbf{0.020675}$ \\
\hline
\end{tabular}
}
\label{table:diff-method}
\end{table*} 

In this section, we conduct experiments on input synthetic images. We use the WHU-RSVI~\cite{cao2020whu} dataset, from which we select two sets of data from trajectory1-fast and trajectory2-fast for 3D reconstruction and pose estimation. We first employ~\cite{purkait2017rolling} to detect RS curves by segmenting curves into multiple short-line segments and performing line fitting for initialization. The GS line-based SfM~\cite{liu20233d} is applied to initialize the \textit{RSL-BA} parameters. The comparative methods include GSBA, NMRSBA, NWRSBA, and GSLBA. Table~\ref{table:diff-method} shows the median absolute trajectory error of different methods, it can be observed that the proposed \textit{RSL-BA} method is the most stable one, achieving optimal or near-optimal results in all cases. Qualitative comparison are also provided in Fig.~\ref{fig: different_ba_result_whu}. Unlike point-based methods, line-based methods often achieve good results with fewer feature lines. However, the GSL-BA method is not sufficiently stable when RS effects are prominent.

\subsection{Real Images}
\label{supp:sec:Real Images}

\begin{figure}[h]
\centering
\includegraphics[width=1\columnwidth]{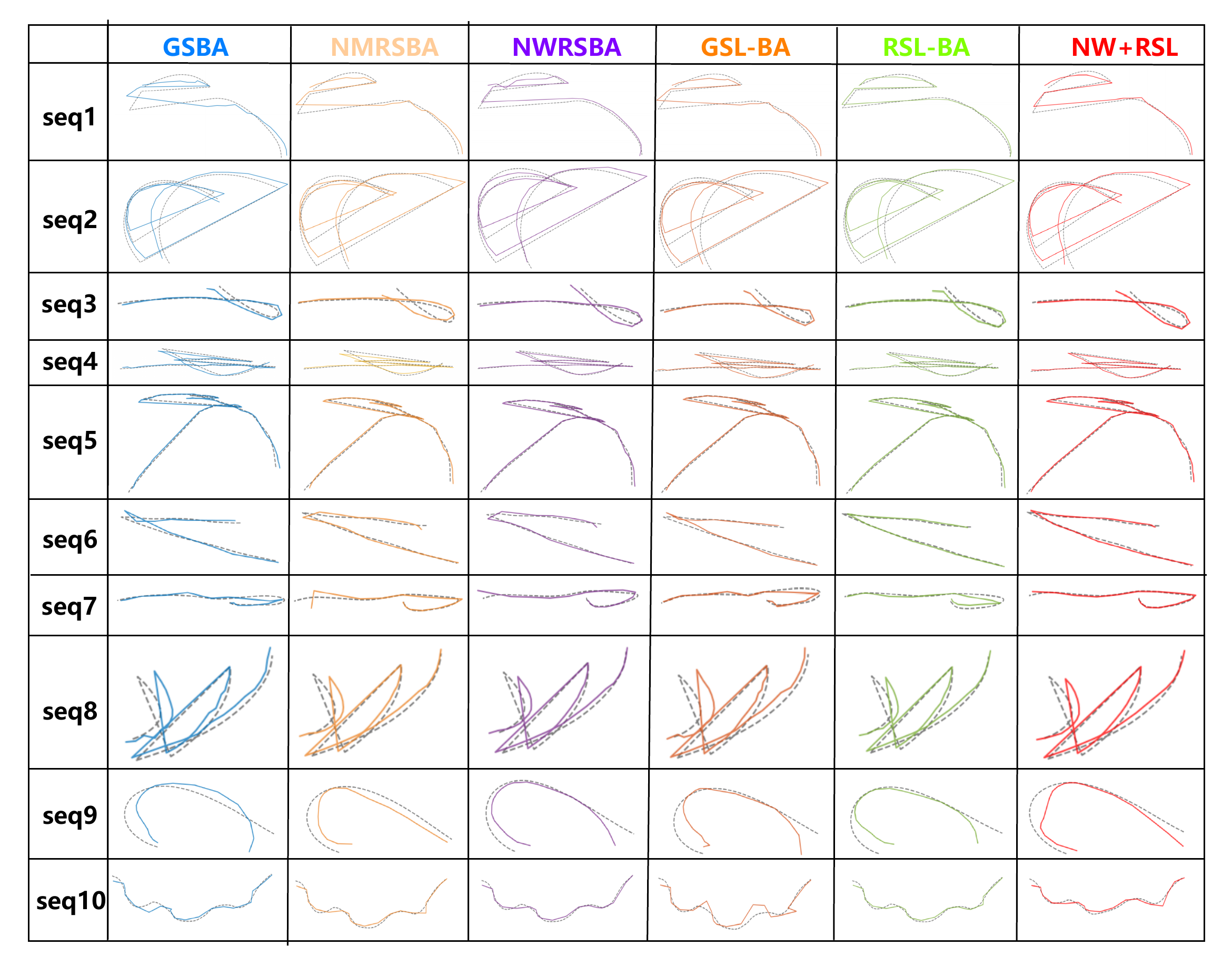}
\caption{Comparison of trajectory errors on the TUM-RSVI\cite{schubert2019rolling}. Each column represents a different bundle adjustment algorithm, and each row represents a different sequence.}
\label{fig: tum_images}
\end{figure}

In this section, we conduct experiments on the real image dataset TUM-RSVI~\cite{schubert2019rolling}. The experimental setup and comparison methods are similar to those in Sec.~\ref{supp: sec: Synthetic Images}. Table~\ref{table:error_tum} shows the median absolute trajectory error of different methods, tracer comparison plots is also provided in Fig~\ref{fig: tum_images}, it can be observed that the RSL-BA method outperforms other methods but is slightly weaker than NWRSBA. This is because the TUM-RSVI lacks line features and they are not visually prominent, making it less suitable for RSL-BA. Besides, we implement a naive point-line RSBA by jointly optimizing points and lines with NWRSBA and proposed RSL-BA. An important observation is that such a naive
NWRSBA+RSL-BA combination significantly outperforms each method individually, which implies that the proposed RSL-BA could be applied solely or combined with the point-based method as point-line BA to the downstream RS vision task such as RSSfM or RSSLAM.

\begin{table*}[h]
\centering
\caption{The median absolute trajectory error (ATE) of different methods in TUM-RSVI\cite{schubert2019rolling} dataset.the best results are highlighted by bold font.}
\resizebox{=0.7\textwidth}{!}{
\begin{tabular}{c  c  c  c  c  c c}
\hline
  & GSBA & GLBA &  NMRSBA & NWRSBA&  RSL-BA(ours) & NWRS+RSL-BA \\
\hline
 seq1 & 0.069484 & 0.086460 & 0.052366 & 0.045064 & 0.037816 & \textbf{0.034495}\\
\hline
 seq2 & 0.029821 & 0.030379 & 0.026028 & \textbf{0.023227} & 0.025132 & 0.024754
\\
 \hline
 seq3 & 0.065160 & 0.063718 & 0.057918 & 0.055001 & 0.048187 &  \textbf{0.045184} \\
 \hline
 seq4 & 0.049613 & 0.052026& 0.032214  & 0.030534 & 0.031903  & \textbf{0.027448} \\
\hline
 seq5 & 0.031860 & 0.035839 & 0.019407 & 0.016066 & 0.017659 & \textbf{0.015068}\\
 \hline
 seq6 & 0.061966 & 0.061792 & 0.032434 & \textbf{0.024658} & 0.026448 & 0.031414 \\
 \hline
 seq7 & 0.051621 & 0.056534 & 0.039154 & 0.039620 & 0.039701 & \textbf{0.033150}\\
 \hline
 seq8 & 0.026403 & 0.028690 & 0.024807 & 0.025926 & 0.024983 & \textbf{0.024486}\\
 \hline
 seq9 & 0.098334 & 0.098212 & 0.082481 & \textbf{0.073580}  & 0.080523 & 0.076613 \\
 \hline
 seq10 & 0.81174  & 0.81180 & 0.59390 & \textbf{0.53296} & 0.57477 & 0.57417\\
\hline
\end{tabular}
}
\label{table:error_tum}
\end{table*}

%
%
\bibliographystyle{splncs04}
\bibliography{supp}